\documentclass{sonic}

\usepackage{titlesec}
\usepackage{microtype}
\usepackage{hyperref}
\usepackage{url}
\usepackage{booktabs}
\usepackage{graphicx}
\usepackage{subcaption}
\usepackage{wrapfig}
\usepackage{amsmath}
\usepackage{enumitem}
\usepackage{algorithm}
\usepackage{algpseudocode}
\usepackage{cleveref}
\usepackage{orcidlink}

\title{SonicSampler: Unified Tile-Aware Kernels for LLM Sampling and Speculative Verification}

\author[*]{\orcidlink{0000-0002-3790-5757} Pragaash Ponnusamy}
\author[*]{\orcidlink{0009-0008-6293-7460} Shivam Sahni}
\author{\orcidlink{0000-0002-6712-1929} Jue Wang}
\author{\orcidlink{0000-0003-2186-8450} Tri Dao}
\affiliation{Together AI}
\contribution[*]{Equal Contribution}

\correspondence{\email{\{pragaash,ssahni,jue,tri\}@together.ai}}
\code{TBA}
\date{May 2026}

\abstract{%
    Sampling in LLM inference comprises a combinatorial set of logit processing, token selection, and verification operations for speculative decoding. However, existing implementations either accelerate only subsets of this pipeline, rely on multiple kernel launches, or assume homogeneous sampling behavior across a batch, limiting support for dynamic serving workloads and preventing efficient CUDA Graph execution.
    
    We present \textbf{SonicSampler}, a unified suite of tile-aware Triton kernels that vertically fuses the complete sampling pipeline into a fixed, workload-aware execution model. Our kernels support dynamic per-request sampling behaviors, including grammar-constrained decoding, repetition, frequency and presence penalties, logit bias, temperature scaling, top-$k$ / top-$p$ / min-$p$ filtering, and speculative verification - within a single batched kernel while remaining fully CUDA Graph-compatible. Central to our approach is a novel hierarchical two-stage top-$k$ algorithm that achieves up to \textbf{10x speedup} over competitive baselines and exploits the low-entropy structure of LLM outputs to enable efficient selection over large vocabularies.
    
    Across heterogeneous speculative decoding workloads, SonicSampler achieves up to \textbf{16x speedup} over state-of-the-art baselines while preserving flexible batched execution.
}

\begin{document}
\maketitle

\section{Introduction}
\label{sec:introduction}

The performance of large language model (LLM) inference is increasingly critical across latency-sensitive applications, from real-time speech interfaces to agentic workflows where smaller models must execute within tight time budgets~\citep{defossez2024moshi,belcak2025small}.
While much of the systems effort has focused on accelerating the model forward pass, the effectiveness of LLM generation also depends on the \textit{efficiency} and \textit{flexibility} of sampling--the process of converting model logits into discrete token decisions.
Sampling is not merely a post-processing detail; rather, it governs output diversity, controllability, and structural validity \citep{holtzman2020curious,zhu2023penalty,willard2023outlines,dong2025xgrammarflexibleefficientstructured}. 
It also enables quality improvement strategies such as best-of-$N$ decoding, where multiple diverse candidates are drawn and selected to enhance the final quality of the output \citep{wang2022self,chen2023universal,li2025selfmoa,wang2025effect}. Improving LLM serving therefore requires not only faster model execution but also a sampling pipeline that is both efficient and expressive under realistic deployment constraints.

This sampling bottleneck is particularly acute in speculative decoding~\citep{leviathan2023fast, chen2023accelerating}, where lightweight draft models propose candidate continuations that must be sampled and verified against the target distribution.
As drafters grow increasingly lightweight, e.g.~Medusa heads~\citep{cai2024medusa}, EAGLE~\citep{li2024eagle}, and multi-token prediction~\citep{gloeckle2024mtp}, the relative cost of the downstream sampling pipeline grows correspondingly. 
In practice, this pipeline is not a single primitive but a heterogeneous composition of operations, including grammar-constrained masking, repetition, frequency and presence penalties, logit bias, temperature scaling, top-$k$ / top-$p$ / min-$p$ filtering, stochastic or greedy token selection, and speculative verification~\citep{fan2018hierarchical,holtzman2020curious,nguyen2024minp,hewitt2022truncation}.

Existing kernel implementations fail to address this complexity adequately. 
Some accelerate isolated components such as grammar masking, penalty application, or probability filtering, but leave the overall pipeline fragmented across multiple kernel launches, incurring additional overhead and intermediate memory traffic.
Others partially fuse sampling, yet assume homogeneous behavior across the batch and, therefore, cannot support the mixed greedy and stochastic configurations that arise under continuous batching \citep{yu2022orca}.
Critically, many top-$k$ implementations are also incompatible with CUDA Graph, forfeiting the launch-overhead amortization that is increasingly essential in production serving engines~\citep{yu2022orca,kwon2023vllm,nvidia2023tensorrtllm,zheng2024sglang}.

In this work, we present \textbf{SonicSampler}, a unified suite of tile-aware Triton~\citep{tillet2019triton} kernels that vertically fuse the complete sampling pipeline, from logit processing through speculative verification, into a CUDA Graph-compatible execution model. Our kernels support both \textit{singular} mode (standard and draft-model inference) and \textit{multistep} mode (verification), with fully dynamic per-request configurations within batched dispatch via compact bit-level indicators. 
Central to our approach is a reformulation of the top-$k$ bottleneck. While filtering nominally requires global reductions across vocabularies of $2^{17}$--$2^{18}$ tokens, practical next-token distributions are highly concentrated; thus, the effective support relevant for sampling is far smaller than the full vocabulary. This motivates a bounded candidate set with $k{=}128$, which is sufficient whenever the probability mass outside the retained set is negligible for the chosen truncation rule. 
We verify this empirically in \Cref{exp:correctness}, showing that top-$128$ retains nearly all relevant mass and preserves downstream accuracy.
 
Leveraging this, we introduce a \textit{two-stage hierarchical top-$k$} algorithm. Stage~1 tiles across vocabulary blocks, applying the full logit-processing prologue before reducing each tile to $k$ candidates via an adaptive radix or bitonic selection, encoded under a monotonicity- and stability-preserving lexicographic scheme. Stage~2 gathers and merges candidates across blocks into the final top-$k$. This map-reduce formulation maximizes arithmetic intensity by fusing compute-dense prologues and epilogues with the reduction stages.
 
As such, we summarize our contributions as follows:
 
\begin{itemize}[leftmargin=2em]
    \item We present a unified CUDA Graph-compatible kernel suite that fuses logit processing, sampling, and speculative verification while supporting mixed greedy and stochastic decoding within a single batched workload.
    \item We introduce a two-stage tile-aware hierarchical top-$k$ algorithm that achieves up to \textbf{10x} speedup over existing radix- and bitonic-based alternatives.
    \item We demonstrate sampling speedups of up to $\mathbf{16x}$ on speculative decoding workloads over FlashInfer and other state-of-the-art baselines.
\end{itemize}

\section{Related Work}
\label{sec:related_work}

Existing sampling systems can be categorized along three axes: 
kernel fusion, support for heterogeneous sampling, and compatibility with CUDA Graph.
While prior work explores different points in this design space, none jointly satisfy all three requirements. 
We provide a focused discussion of representative approaches here, 
and defer a more comprehensive review to \Cref{app:related} 
and present a feature comparison in \Cref{app:feature_comparison}.

\paragraph{Standalone sampling kernels.}
Kernel libraries such as FlashInfer~\citep{flashinfer2024} and XGrammar~\citep{dong2025xgrammarflexibleefficientstructured} provide efficient implementations of key components in the sampling pipeline, including top-$k$/top-$p$ selection and grammar-constrained decoding. These works establish strong building blocks for production systems, though they operate as separate kernels when composed together.

\paragraph{Fused sampling kernels.}
Some recent approaches \citep{modular2025max} introduce partial fusion of sampling within a single kernel. However, these designs assume homogeneous sampling configurations and cannot natively support mixed greedy and stochastic decoding within a batched workload, limiting their applicability in production serving.

Orthogonal efforts such as MegaKernel~\citep{cheng2025miragepersistentkernelcompiler} fuse the model forward pass into a single execution unit, reducing kernel launch overhead, and are complementary as they target model execution rather than sampling.

FlashSampling~\citep{ruiz2026flashsampling} reduces memory traffic by fusing the LM head with downstream sampling, targeting an inference stage complementary to post-logit processing. It also briefly notes a two-stage decomposition of top-$k$ for future work. Our work instead focuses on the system design required to realize such a decomposition efficiently in practice.

\paragraph{Efficient top-$k$ selection.}
A separate line of work focuses on accelerating top-$k$ selection itself, spanning multi-pass radix selection~\citep{radik2024,wang2025tilelangcomposabletiledprogramming}, bitonic-based streaming kernels~\citep{tillet2019triton}, and hybrid or distribution-aware schemes~\citep{flashinfer2024,park2026qritahighperformancetopktopp}. 
These methods treat selection as an isolated primitive and do not account for its tight coupling with upstream logit transformations and downstream sampling or verification steps, missing opportunities for vertical fusion.

Taken together, these works provide efficient primitives and partial fusion strategies for LLM inference and sampling. SonicSampler builds on these insights by unifying logit processing, sampling, and verification within a CUDA Graph-compatible execution model that supports heterogeneous batched workloads.

\section{SonicSampler} \label{sec:method}

This section presents SonicSampler in detail. Our design is guided by three principles:
(1) exploit the inherent structure of the sampling workload to maximize fusion opportunities,
(2) provide theoretical grounding for algorithmic simplifications, 
and (3) support the full heterogeneity of production serving within a single batched dispatch.


\subsection{Background}
\label{sec:background}

To ground the design of SonicSampler, we first introduce the LLM sampling pipeline and its constituent operations. These operations define the computational structure that our execution model seeks to optimize.

LLM sampling transforms model logits into discrete tokens through a sequence of logit processing, truncation, sampling, and (optionally) speculative verification steps. Given logits $\mathbf{x} \in \mathbb{R}^V$, these operations produce a probability distribution and select the next token.

\paragraph{Logit Processing.}
We view logit processing as a sequence of transformations applied to the raw logits, 
including grammar masking, repetition and frequency penalties, logit bias, and temperature scaling~\citep{keskar2019ctrl,dong2025xgrammarflexibleefficientstructured,hinton2015distilling}.
These operations act independently on each token and, therefore, form a \textit{map-style} computation that is fully parallelizable across the vocabulary.

\paragraph{Probability Truncation.}
Truncation methods such as top-$k$, top-$p$, and min-$p$ select a subset of candidate tokens~\citep{fan2018hierarchical,holtzman2020curious,nguyen2024minp}.
In contrast to logit processing, these operations depend on comparisons across tokens (e.g., ranking or cumulative probability) 
and thus require \textit{global reduction} over the vocabulary. 
Notably, we show in \cref{eq:min_p_trick} (Appendix) that min-$p$ can be reformulated to avoid explicit softmax.

\paragraph{Top-$k$ Selection Primitives.}
Ranking-based truncation on GPUs is typically realized through general-purpose top-$k$ selection primitives.
\emph{Radix-based selection} partitions keys by successive digit groups from most to least significant, recursing into the bucket containing the $k$-th element at each pass~\citep{alabi2012fast}.
\emph{Bitonic-based selection} builds on bitonic sorting networks~\citep{batcher1968sorting}, whose regular, branch-free structure maps naturally onto SIMT execution and especially effective for tile-local, fixed-size reductions~\citep{satish2009designing,shanbhag2018efficient}.


\paragraph{Sampling.}
Token selection can be expressed as an $\arg\max$ over perturbed logits via Gumbel-Max reparameterization~\citep{jang2016categorical}, unifying stochastic and greedy sampling under a single formulation: $x^* = \arg\max_i (x_i + \epsilon_i)$, where $\epsilon_i$ introduces stochasticity.

\paragraph{Speculative Verification.}
Speculative decoding accelerates inference by having a draft model propose candidate tokens, which are accepted when $u \cdot q(d) \leq p(d)$ for $u \sim \mathcal{U}(0,1)$, or, in the case of a rejection, sampled from the residual distribution $\tilde{p}(x) \propto \max\{0,\, p(x) - q(x)\}$ \citep{leviathan2023fast}.

These components exhibit a characteristic structure, i.e.~element-wise transformations followed by global reductions, 
which motivates the workload-aware execution model introduced next. 
Detailed formulations are deferred to \Cref{app:background}.

\begin{figure*}[t]
  \centering
  \includegraphics[width=\linewidth]{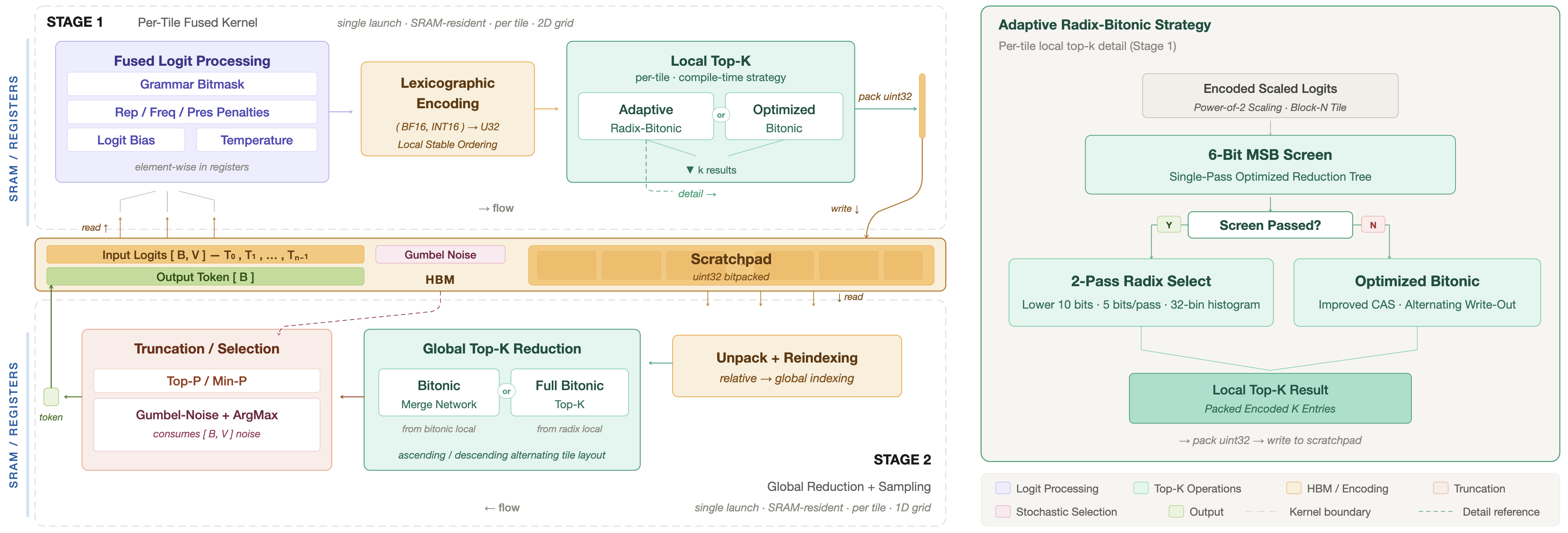}
  \caption{\small SonicSampler's two-stage tile-aware fused sampling pipeline for the Singular (\textit{non-speculative}) workload. 
  \textbf{Stage 1} (\textit{top, left-to-right}) vertically fuses grammar masking, repetition penalties, logit bias, and temperature scaling with Lexicographic Encoding and per-tile Local Top-$k$ via Optimized Bitonic or Adaptive Radix-Bitonic selection (\textit{right panel}), writing only the packed $k$ results per tile to the HBM scratchpad as the sole cross-kernel memory boundary. 
  \textbf{Stage 2} (\textit{bottom, right-to-left}) re-indexes across tiles, performs global Top-$k$ reduction through bitonic merge networks with ascending / descending alternating tile layout, applies probability truncation, and emits the selected token via Gumbel-Noise and ArgMax, completing a clockwise data loop through the shared HBM band. 
  The Adaptive Radix-Bitonic strategy (\textit{right}) screens on the 6 most-significant bits via a single-pass optimized reduction tree, branching to a 2-Pass Radix Select on the lower 10 bits or falling back to an Optimized Bitonic selection.}
  \label{fig:center-spread}
\end{figure*}


\subsection{Workload-Aware Execution Model}
\label{sec:workload_aware}

Existing implementations dispatch each sampling stage as a separate kernel, incurring launch overhead and intermediate materialization, while the top-$k$ reduction itself executes as a single-program streaming pass over the full vocabulary, thereby limiting occupancy to $B$ concurrent programs and precluding tiled execution. We decompose the global reduction into a \emph{two-stage hierarchical} operation over vocabulary tiles of size $B_N$:
\begin{enumerate}[leftmargin=*,itemsep=2pt,topsep=2pt]
    \item \textbf{Stage 1} launches $B \cdot Z_v$ programs (where $Z_v = \left\lceil \frac{V} {B_N} \right\rceil$), each fusing the logit-processing prologue with a tile-local top-$k$ reduction that emits $k$ packed candidates to a scratchpad in global memory.
    \item \textbf{Stage 2} launches $B$ programs, each gathering the $Z_v \cdot k$ packed candidates and performing a cross-tile merge followed by a vertically fused epilogue of probability truncation, Gumbel perturbation, and $\arg\max$ selection.
\end{enumerate}
This decomposition increases Stage~1 parallelism by a factor of $Z_v$ relative to single-stage formulations while confining all inter-tile communication to a compact $B \cdot Z_v \cdot k$ scratchpad of packed \texttt{uint32} entries.
The fusion boundaries are chosen so that no intermediate logit or probability vector is ever materialized at vocabulary scale between the two stages.


\subsection{Entropy Sufficiency for Bounded Top-$k$}
\label{sec:entropy_sufficiency}

The two-stage decomposition requires bounding the number of candidates retained from each tile. We establish that for well-trained LLMs, a modest bound of $k = 128$ suffices to capture the effective probability mass under the practical top-$p$ threshold.
  
\paragraph{Theoretical Analysis.} Consider a probability distribution over $V$ tokens where mass $P$ is concentrated in $K$ candidates (the top-$k$ set) and mass $(1-P)$ is distributed over the remaining $V - K$ tokens. The Shannon entropy of such a distribution is bounded by:
\begin{equation}
    H \leq H^{(2)}(P) + P \log_2 K + (1 - P) \log_2 (V - K)
\end{equation}
where $H^{(2)}(P) = -P \log_2 P - (1-P) \log_2 (1-P)$ is the binary entropy. This follows from the Schur-concavity of entropy, where the uniform distribution of mass within each partition maximizes entropy. Rearranging for $K$ when $V \gg K$, we'd have:
\begin{equation}
    K \geq \frac{P}{V^{(1/P) - 1}} \cdot 2^{H/P} = PV \cdot \sqrt[P]{\frac{\text{Perplexity}}{V}}
    \label{eq:k_bound}
\end{equation}
For $K = 128$ and $V = 2^{17}$, the maximum satisfiable entropy is therefore given by:
\begin{equation}
    H \leq 17 - P(10 + \log_2 P)
\end{equation}
Now, with $P = 0.95$ (a typical top-$p$ threshold), this yields $H \lesssim 7.6$ bits. In the limiting case of $\tau \to \infty$ where $H \to \log_2 V$, we would require $K \geq PV$, but this regime is practically irrelevant.

\paragraph{Empirical Grounding.} \citet{du2025temperature} show that optimal temperatures for multi-sample inference typically satisfy $\tau < 1.0$, where higher temperatures increase improper token selection. 
Consequently, well-trained LLMs produce peaked distributions with probability mass concentrated in a small candidate set.
This places practical sampling in the regime $H \leq 7$ bits, making $k = 128$ sufficient. We validate this in \Cref{exp:correctness}, showing negligible residual mass beyond top-$k$ and no loss in downstream accuracy on GPQA-Diamond.


\subsection{Two-Stage Hierarchical Top-$k$}
\label{sec:two_stage}

We decompose vocabulary-scale top-$k$ selection into two stages: tile-local reduction in the first stage, followed by cross-tile merging in the second.
This decomposition maximizes parallelism by launching independent programs across vocabulary tiles, each fusing the logit processing prologue with local candidate selection.
The tile-local reduction admits two strategies, namely an optimized bitonic network and an adaptive radix-bitonic filter (see \Cref{fig:center-spread}), which we detail after establishing the encoding that underlies both.


\paragraph{Numerical Lexicographic Encoding.} To enable correct ordering across the sign boundary while preserving index stability, we encode \texttt{bfloat16} values into a lexicographically ordered unsigned integer representation. Given $x \in \mathbb{F}_{16}$, let $\beta(x)$ denote its bit-reinterpretation as \texttt{uint16} and $\sigma(x) = \mathbf{1}[\beta(x) \geq 2^{15}]$ be the sign indicator. The encoding is thus given by:
\begin{equation*}
    \phi(x) = \beta(x) \oplus \mu(\sigma(x)) \hspace{0.7cm} \text{s.t.} \hspace{0.7cm} \mu(s) = \begin{cases}
        2^{15}, & \text{if } s = 0 \\
        2^{16} - 1, & \text{otherwise}
    \end{cases}
\end{equation*}
This maps non-negative values to $[2^{15}, 2^{16}-1]$ via MSB toggle (preserving order) and negative values to $[0, 2^{15}-1]$ via bitwise complement (reversing order to maintain monotonicity with increasing magnitude). The encoding is self-dual:
\begin{equation*}
    \phi^{-1}(y) = y \oplus \mu(1 - \sigma'(y)) \hspace{0.7cm} \text{s.t.} \hspace{0.7cm} \sigma'(y) = \mathbf{1}[y \geq 2^{15}]
\end{equation*}
For stable sorting, we pack the encoded values in the upper-16 bits with the inverted indices in the lower half, i.e. the finalized transform being given by:
\begin{equation*}
    \psi(x, i, n) = \phi(x) \cdot 2^{16} + (n - 1) \oplus i
\end{equation*}
where $n$ is the tile size and $i \in \{0, \ldots, n-1\}$ is the position index. The index complement ensures descending index priority among tied values, guaranteeing stability. We also generalize this transform to enable alternating relative indices, detailed in \Cref{app:alternating_indices}.


\paragraph{Optimized Bitonic Top-$k$ Selection.}
Our primary reduction employs bitonic networks for their regular, data-independent, and highly parallelizable comparison patterns.
\Cref{alg:cas,alg:bitonic_fold} present the critical pieces of our innnovation while supplementary and encapsulating portions of the overall algorithm (\textsc{HypercubeMerge}, \textsc{BitonicSelection}) are detailed in \Cref{app:pseudocode}. 
The \textsc{CAS} primitive (\Cref{alg:cas}) expresses the conditional swap as explicit $\min$/$\max$ operations followed by a conditional select, rather than the conventional bitwise-XOR formulation. This enables the compiler to emit fused \texttt{IMNMX.U32.U32} instructions that perform each compare-and-swap in a single cycle on SM90+ hardware.
 
 \textsc{BitonicTopK} (\Cref{alg:bitonic_topk}) operates on a hypercube view of the packed tile in two phases: $\kappa = \log_2 k$ merge stages sort every $k$-width slice, then a $\max$-reduction along the fold axis reduces the tile from $B_N$ to $R \cdot k$ candidates within the register file.
The remaining $F$ folds are performed by \textsc{BitonicFold} (\Cref{alg:bitonic_fold}), which, at each iteration, reshapes into $[R/2^i,\, 2,\, k]$ and transposes to $[R/2^i,\, k,\, 2]$ before splitting.
This reshape-transpose-split sequence forces the compiler to emit a \texttt{convert\_layout} that places a complete pair of $k$-element sub-tiles within each warp, compiling the subsequent $\max$ to a warp-local \texttt{IMNMX} reduction tree with no inter-warp barriers. While this sequence inevitably triggers an exchange through shared memory, i.e., a \texttt{STS-BAR-LDS} chain, this re-layout is effectively only triggered once, while subsequent sequences share the same layout and are compiled away.
The order parameter $\delta = \texttt{block\_id} \bmod 2$ alternates sort direction across tiles, enabling Stage~2 to merge pre-sorted alternating sub-tiles via \textsc{BitonicReduction} (\Cref{alg:bitonic_reduction}) without re-sorting.

\begin{figure}[t]
\begin{minipage}[t]{0.48\linewidth}
\begin{algorithm}[H]
\caption{Compare-and-Swap}\label{alg:cas}
\small
\begin{algorithmic}[1]
\Require Packed $\mathbf{v} \in \mathbb{N}^{2^n}$ (hypercube), pair $\rho \in \{0,1\}^2$, flip $f$, dimension $d$, total
$n$
\Ensure Swapped $\mathbf{v}' \in \mathbb{N}^{2^n}$
\State $o \gets n - d - 1$ \Comment{Outer axis}
\State $\mathbf{r} \gets \mathbf{v} \oplus \textsc{XorReduce}(\mathbf{v}, \text{axis}=o)$
\State $\sigma \gets f \oplus \textsc{Expand}(\rho, o, d)$
\State \Return $\textsc{Select}(\sigma \neq 0,\ \max(\mathbf{v}, \mathbf{r}),\ \min(\mathbf{v}, \mathbf{r}))$
\end{algorithmic}
\end{algorithm}
\end{minipage}
\hfill
\begin{minipage}[t]{0.48\linewidth}
\begin{algorithm}[H]
\caption{Screen-Combine}\label{alg:screen_combine}
\small
\begin{algorithmic}[1]
\Require $(m_a, c_a), (m_b, c_b) \in \mathbb{N} \times \mathbb{N}$
\Ensure $(m', c') \in \mathbb{N} \times \mathbb{N}$
\State $m' \gets \max(m_a,\, m_b)$ \Comment{IMNMX}
\State $\hat{c} \gets \textsc{Select}(m_a {=} m_b,\, c_a + c_b,\, c_b)$ \Comment{IADD}
\State $c' \gets \textsc{Select}(m_a {>} m_b,\, c_a,\, \hat{c})$
\State \Return $(m', c')$
\end{algorithmic}
\end{algorithm}
\end{minipage}
\end{figure}


\paragraph{Adaptive Radix-Bitonic Selection.}
As an alternative, we exploit a distributional property of LLM logits: ${\sim}99.82\%$ of vocabulary tiles share a single mode in the upper 6 bits of their $\phi$-encoded values at unit scale, rising to $99.99\%$ with a power-of-two scaling factor (e.g., $s=4$).
This permits radix-based threshold computation over only the lower 10 bits.
 
\Cref{alg:radix_bitonic} describes the strategy.
The screening phase (lines~1--4) scales, encodes, and performs a single fused reduction via \textsc{ScreenCombine} (\Cref{alg:screen_combine}) that simultaneously computes the modal 6-bit value $\theta$ and its frequency $c$.
The combine operator tracks both quantities in one tree reduction, compiling to an interleaved \texttt{IMNMX}--\texttt{IADD} sequence that avoids a separate counting pass and halves inter-warp synchronization relative to a two-pass approach.
 
When $c \geq k$ (sparse path), \textsc{SparseThreshold} (\Cref{alg:sparse_threshold}) performs two 5-bit radix passes over 32-bin warp-local histograms.
Each pass invokes \textsc{RadixPartition} (\Cref{alg:radix_partition}), which identifies the pivot via a sum-based count (\texttt{IADD} tree) and computes the margin via a masked-max (intra-warp \texttt{IMNMX} tree).
Input values are XOR-inverted before histogramming, transforming the suffix-sum needed for descending top-$k$ into an ascending prefix-sum compatible with the hardware \textsc{CumSum} primitive.
When $c < k$ (dense path), the algorithm falls back to the optimized bitonic top-$k$ (\Cref{alg:bitonic_selection}).

\begin{figure}[t]

    
    \begin{algorithm}[H]
        \caption{Bitonic Fold}\label{alg:bitonic_fold}
        \small
        \begin{algorithmic}[1]
        \Require Block $\mathbf{G} \in \mathbb{N}^{R \times k}$, pair $\rho$, rows $R$, target $k$, folds $F$, order $\delta$
        \Ensure Reduced candidates $\mathbf{y} \in \mathbb{N}^k$
        \State $n \gets \log_2(R \cdot k) + 1$,\quad $\kappa \gets \log_2 k$
        \State $\mathbf{B} \gets \mathbf{G}$
        \For{$i \gets 1$ to $F$}
            \State $\mathbf{B} \gets \textsc{Reshape}(\mathbf{B},\ [R / 2^i,\ 2,\ k])$
                \Comment{$\triangleright$ Pair adjacent sub-tiles along dim 1}
            \State $\mathbf{L}, \mathbf{R} \gets \textsc{Split}(\textsc{Transpose}(\mathbf{B},\ [0,2,1]))$
                \Comment{$\triangleright$ Triggers \texttt{convert\_layout}: each warp holds a full $(\mathbf{L}, \mathbf{R})$ pair}
            \State $\mathbf{B} \gets \max(\mathbf{L},\ \mathbf{R})$
                \Comment{$\triangleright$ Warp-local \texttt{IMNMX} tree, zero inter-warp \texttt{bar.sync}}
            \State $\hat{\mathbf{H}} \gets \textsc{Reshape}(\mathbf{B},\ [2]^{n - i - 1})$
                \Comment{$\triangleright$ Hypercube view for bitonic merge}
            \If{$i < F$}
                \State $f \gets \textsc{Expand}(\rho,\ F{-}i{-}1,\ \kappa)$
                    \Comment{$\triangleright$ Intermediate fold: parity-dependent ordering}
                \State $\hat{\mathbf{H}} \gets \textsc{HypercubeMerge}(\hat{\mathbf{H}}, \rho, f, \kappa, n{-}i{-}1)$
            \Else
                \State $\hat{\mathbf{H}} \gets \textsc{HypercubeMerge}(\hat{\mathbf{H}}, \rho, \delta, \kappa, n{-}i{-}1)$
                    \Comment{$\triangleright$ Final fold: caller-specified order $\delta$}
            \EndIf
            \State $\mathbf{B} \gets \hat{\mathbf{H}}$
        \EndFor
        \State \Return $\textsc{Reshape}(\mathbf{B},\ [k])$
            \Comment{$\triangleright$ Flatten collapsed hypercube to $k$ winners}
        \end{algorithmic}
    \end{algorithm}


    \begin{algorithm}[H]
        \caption{Adaptive Radix-Bitonic Selection}\label{alg:radix_bitonic}
        \small
        \begin{algorithmic}[1]
        \Require Logits $\mathbf{x} \in \mathbb{R}^{B_N}$, indices $\mathbf{I}$, target $k$, scale $s$, \textsc{block\_id}
        \Ensure Top-$k$ candidates written to scratchpad $Y$
        \State $\mathbf{e} \gets \phi(\mathbf{x} \cdot s)$ \Comment{Encode scaled logits (\texttt{uint16})}
        \State $\mathbf{u} \gets \mathbf{e} \gg 10$ \Comment{Upper 6 bits}
        \State $(\theta, c) \gets \textsc{Reduce}\bigl((\mathbf{u},\, \mathbf{1}),\ \textsc{ScreenCombine}\bigr)$
          \Comment{Fused max-count scan}
        \State $\mathbf{M} \gets (\mathbf{u} = \theta)$
        \If{$c \geq k$} \Comment{Sparse path: radix over lower 10 bits}
          \State $\theta, m \gets \textsc{SparseThreshold}(\mathbf{e},\, \mathbf{M},\, \theta,\, k)$
              \Comment{Two-pass 5-bit radix (\Cref{alg:sparse_threshold})}
          \State $\textsc{RadixSelection}(Y,\, \mathbf{e},\, \theta,\, m,\, \mathbf{I},\, k,\, \textsc{block\_id})$
        \Else \Comment{Dense path: bitonic fallback}
          \State $\mathbf{p} \gets \psi(\mathbf{e},\, \mathbf{I},\, B_N)$ \Comment{Pack with inverted indices}
          \State $\textsc{BitonicSelection}(Y,\, \mathbf{p},\, k,\, \textsc{block\_id})$
        \EndIf
        \end{algorithmic}
    \end{algorithm}
    
    
\end{figure}


\paragraph{Stage 1: Tile-Local Reduction.}
The first-stage kernel tiles across batch rows and vocabulary blocks.
Each program instance loads a $B_N$-element logit tile, applies the logit-processing prologue, i.e., grammar masking, repetition, frequency, and presence penalties, logit bias, and temperature scaling, conditioned on per-request bit-level indicators (\Cref{sec:bit_indicators}), and invokes either the bitonic or adaptive radix-bitonic reduction.
The bitonic path writes $k$ packed candidates in alternating ascending-descending order keyed on \texttt{block\_id} parity, while the adaptive path writes $k$ candidates in coalesced order via a \texttt{cumsum-based} scatter. Full pseudocode is provided in \Cref{alg:stage1,alg:stage2} within \Cref{app:pseudocode}.
 
\paragraph{Stage 2: Cross-Tile Merge.}
The second-stage kernel gathers $Z_v \times k$ packed entries from the scratchpad, remaps tile-local indices to absolute vocabulary positions, and performs a final cross-tile top-$k$ reduction.
When Stage~1 uses the bitonic path, the alternating sort order permits a direct \textsc{BitonicReduction} (\Cref{alg:bitonic_reduction}) that merges pre-sorted sub-tiles without re-sorting; otherwise, a full \textsc{BitonicTopK} is applied.
Following the merge, the kernel decodes the packed entries, applies the probability-truncation epilogue (top-$k$, top-$p$, min-$p$ masking via cumulative softmax), adds pre-drawn Gumbel noise, and selects the final token via $\arg\max$.
For speculative verification, it additionally materializes the draft probabilities over the selected indices via fast, stable softmax computation.

\paragraph{Complexity Analysis.}
Let $V$ denote vocabulary size, $Z_v = \left\lceil \frac{V} {B_N} \right\rceil$ the number of tiles, $k$ the candidate bound, and $\kappa = \log_2 k$. Stage~1 launches $B \cdot Z_v$ programs.

On the \emph{bitonic path}, each program performs $O(B_N \log^2 k)$ work: the initial $\kappa$ merge stages contribute $\sum_{s=1}^{\kappa} s$ CAS passes over $B_N$ elements, i.e., $O(B_N \cdot \kappa^2)$, dominating the subsequent fold phase with complexity $O(B_N \cdot \kappa)$.

On the \emph{radix path} (sparse case), each program performs $O(B_N)$ work via a single screening reduction and two histogram passes.
Stage~2 launches $B$ programs: $O(Z_v \cdot k \cdot \log k)$ per program when merging alternating sub-tiles (fold-only reduction), or $O(Z_v \cdot k \cdot \log^2 k)$ when applying a full bitonic top-$k$ to unordered sub-tiles.

Since $Z_v \cdot k \ll V$, Stage~1 dominates in both cases, yielding $O(B \cdot V \cdot \log^2 k)$ total work for the bitonic path and $O(B \cdot V)$ for the radix path.
A single-program streaming top-$k$ performs $O(V \cdot \log k)$ work per row but admits only $B$ concurrent programs; our formulation trades a modest work factor of $\log k$ (bitonic) or none (radix) for a $Z_v{\times}$ increase in Stage~1 parallelism.
 

\subsection{Batch-Level Dynamism via Bit-Level Indicators}
\label{sec:bit_indicators}

\begin{wrapfigure}{r}{0.35\linewidth}
    \vspace{-2.5em}
    \centering
    \begin{tabular}{@{}cl@{}}
    \toprule
    \textbf{Bit(s)} & \textbf{Operation} \\
    \midrule
    \texttt{0x001} & Grammar bit-mask \\
    \texttt{0x002} & Repetition penalty \\ 
    \texttt{0x004} & Frequency penalty \\
    \texttt{0x008} & Presence penalty \\
    \texttt{0x010} & Logit bias \\
    \texttt{0x020} & Temperature scaling \\
    \texttt{0x040} & Greedy selection \\
    \texttt{0x080} & Top-$k$ filtering \\
    \texttt{0x100} & Top-$p$ filtering \\
    \texttt{0x200} & Min-$p$ filtering \\
    \texttt{0x400} & Top-$k$ log-probabilities \\
    \bottomrule
    \end{tabular}
    \caption{Bit-level encoding of sampling operations in SonicSampler.}
    \label{tab:bitcode}
    \vspace{-1em}
\end{wrapfigure}

Production serving requires heterogeneous mixtures of greedy and stochastic sampling within a single batch.
Instead of host-side conditioning and multiple kernel dispatches, which break CUDA Graph compatibility, 
SonicSampler encodes each request as a compact bit-level indicator (see \Cref{tab:bitcode}) loaded once per program instance. 
Consequently, the subsequent execution paths and parameter loads are conditioned on this indicator, allowing greedy and stochastic requests to coexist within a single batched launch without host-side branching. 

Notably, greedy requests follow a streamlined series of local \texttt{argmax} paths, while stochastic requests execute the full pipeline. On NVIDIA Hopper and newer GPUs, we further leverage Programmatic Dependent Launch (PDL) to overlap the two stages by triggering Stage 2 preparation as Stage 1 completes, effectively hiding Stage 2 launch latency at larger batch sizes.

\section{Experiments}
\label{sec:experiments}

We evaluate SonicSampler across three key dimensions:
\begin{enumerate}
    \item \textit{End-to-end decoding performance} under realistic, heterogeneous workloads,
    \item \textit{Kernel-level efficiency} of the sampling pipeline, with a focus on our two-stage top-$k$ algorithm, and
    \item \textit{Correctness}, ensuring that bounded top-$k$ sampling preserves distributional fidelity and downstream task accuracy.
\end{enumerate}

\subsection{Experimental Setup} \label{sec:setup}

\begin{wrapfigure}{r}{0.4\linewidth}
  \vspace{-1.8em}
  \centering
  \includegraphics[width=\linewidth]{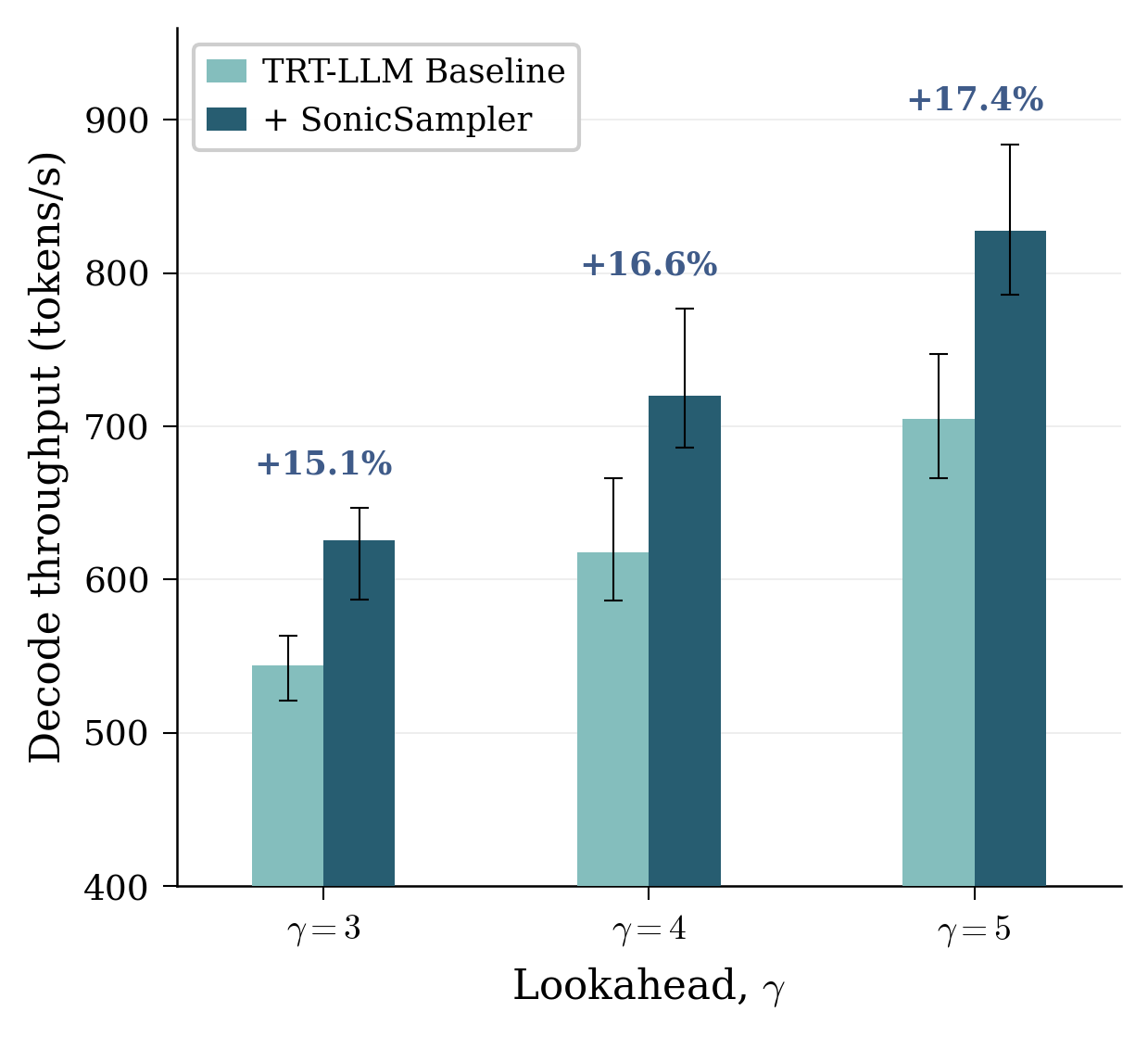}
  \captionof{figure}{\small End-to-end decode throughput with Eagle3 on Qwen3-8B TP=4 with T=0.6, top-$p$=0.9,
top-$k$=128.}
  \label{fig:e2e}
  \vspace{-2em}
\end{wrapfigure}

Experiments are run on NVIDIA B200 GPUs (driver 575.57.08) with Triton v3.5.1.
All floating-point inputs, e.g., logits, use \texttt{bfloat16}, while integer-based inputs use \texttt{int32}. Latency is measured with \texttt{do\_bench\_cudagraph} from the \texttt{triton.testing} suite, with a 500 ms repetition time per configuration before taking the mean latency, synchronizing the GPU, flushing the CUDA cache, and invoking garbage collection between iterations.

\subsection{End-to-End Performance}

\begin{figure*}[t]
    \centering
    \vspace{-1em}
    \includegraphics[width=\linewidth, trim=0 0 0 0.95cm, clip]{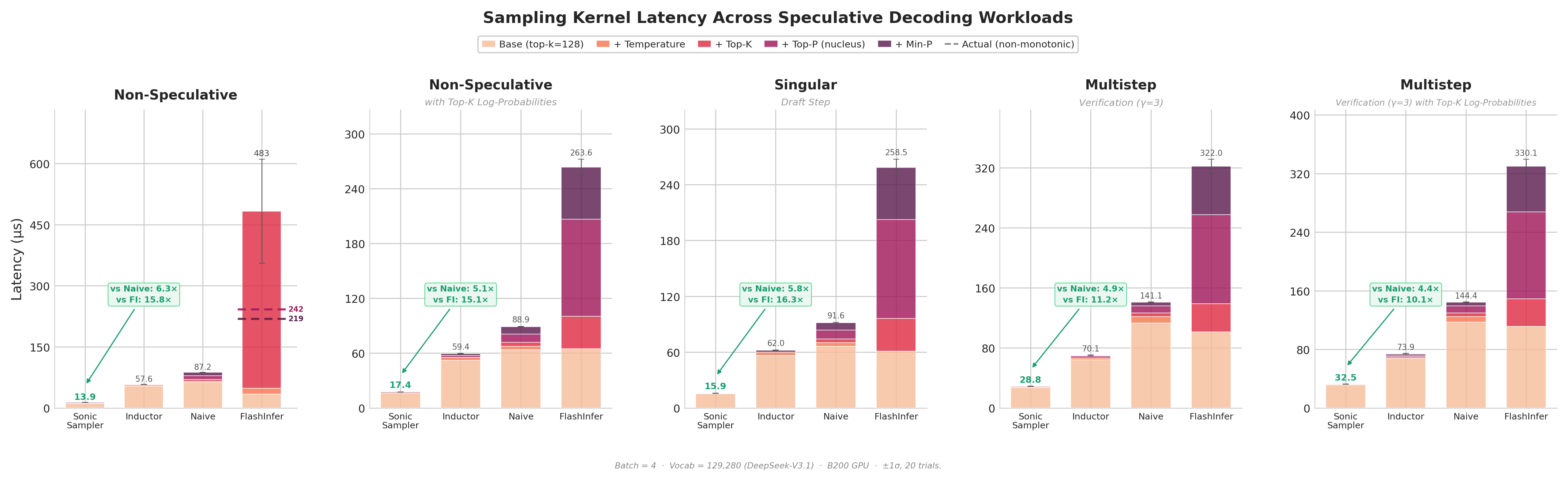}
    \caption{\small Sampling latency breakdown by implementation across workloads with each panel corresponding to a decoding regime. Stacked segments show the incremental cost of each sampling modifier. The dashed lines for FlashInfer indicate that distribution truncation via top-p and min-p reduces pivot overhead, yielding non-monotonic scaling.}
    \label{fig:workload_stacked}
\end{figure*}

We evaluate SonicSampler in end-to-end decoding to measure its impact on practical inference performance. 
Experiments are conducted on Qwen3-8B with Eagle3 using stochastic sampling ($\tau = 0.6$, $\text{top-}p = 0.9$, $\text{top-}k = 128$). 
Since TRT-LLM does not natively sample during drafting, we extend it with its own target-model sampling kernels for a fair comparison.

As shown in \Cref{fig:e2e}, SonicSampler improves decoding throughput across all lookahead values, achieving 15-17\% speedups (+80-120 TPS) over the TRT-LLM baseline. The advantage grows with $\gamma$, reflecting the increasing share of sampling in per-token latency under speculative decoding. By unifying the sampling pipeline into a single CUDA Graph-compatible execution, SonicSampler reduces kernel launch overhead and intermediate memory traffic, translating directly into system-level speedups.

\subsection{Sampling Latency Breakdown}

\Cref{fig:workload_stacked} decomposes sampling latency across five decoding regimes: non-speculative decoding ($\pm$ top-$k$ logprob), singular draft step, and multi-step verification ($\pm$ top-$k$ logprob). Within each regime, we progressively enable sampling operations, from base sampling to temperature scaling, top-$k$, top-$p$, and min-$p$ filtering. We compare against FlashInfer and two additional baselines: \textit{Naive}, which mirrors SonicSampler's pipeline using native PyTorch operations without dynamic workload support, and \textit{Indicator}, its \texttt{torch.compile}d variant.

SonicSampler consistently achieves the lowest latency across all regimes, yielding approximately 2.5--4$\times$ over Indicator, 5--6$\times$ over Naive, and 10--16$\times$ over FlashInfer. 
Baseline implementations incur progressively higher costs due to repeated kernel launches and intermediate memory traffic, leading to steadily increasing latency.
In particular, FlashInfer exhibits a pronounced increase in latency when top-$k$ is applied. In contrast, SonicSampler's fused design absorbs these operations with minimal marginal overhead.

\begin{figure}[t]
    \centering
    \vspace{-1em}
    \includegraphics[width=\linewidth, trim=0 0 0 1.75cm, clip]{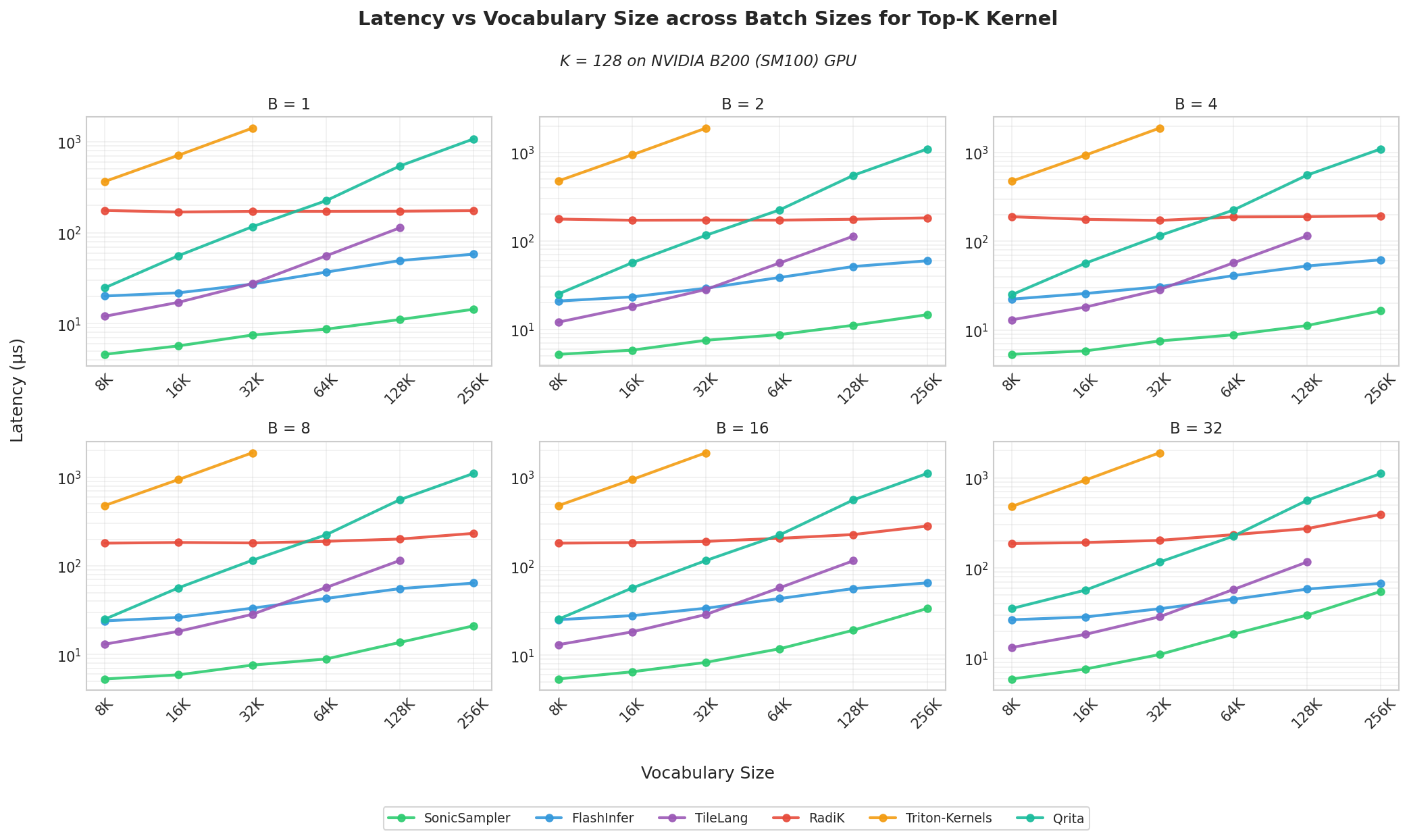}
    \caption{\small Top-$k$ kernel latency ($\mu$s, log scale) vs.\ vocabulary size $V$ for $k = 128$ across batch sizes $B \in \{1, 2, 4, 8, 16, 32\}$.
    Triton-TopK is limited to $V \leq 2^{15}$; TileLang-TopK to $V \leq 2^{17}$.}
    \label{fig:topk-latency}
\end{figure}

\subsection{Top-$k$ Kernel Performance}

We evaluate Top-$k$ selection in isolation to quantify the efficiency of our two-stage algorithm for large-vocabulary sampling.
We use 2D logit tensors of shape $(B, V)$ with fixed $K=128$, sweeping $V$ from $2^{13}$ to $2^{18}$ and $B \in \{1,2,4,8,16,32\}$. 
Latency is measured on inputs drawn from Zipf($\alpha=1.1$) following prior work~\citep{radik2024, park2026qritahighperformancetopktopp}; 
correctness is additionally verified on $\mathcal{U}(128.6,128.7)$ and all-zero inputs. 
All baselines are run under matched configurations (precision, input shapes, and hardware), e.g., TileLang uses its native compiler, while RadiK is evaluated as a precompiled CUDA C++ binary. For a subset of methods, we apply minimal adaptations to ensure correctness and compatibility 
with large vocabulary sizes and \texttt{bfloat16} inputs. Notably, these modifications are conservative and, where measurable, do not degrade and, in some cases, slightly improve the baseline performance.

\paragraph{FlashInfer.}
FlashInfer's sampling kernel generates Gumbel noise internally from a random seed, which breaks CUDA graph compatibility. 
We modify the kernel to accept precomputed Gumbel noise as an external input, 
enabling CUDA graph capture and ensuring a fair comparison with our approach.
We also adapt the kernel to operate on \texttt{bfloat16} inputs to match our evaluation precision.

\paragraph{TileLang-TopK.}
The original implementation supports only \texttt{float32} inputs. We extend it to support \texttt{bfloat16} by introducing an explicit cast path during input loading and promoting values to \texttt{float32} for intermediate computation.
We additionally increase the shared-memory buffer size from 4,096 to 16,384 entries to avoid silent truncation of candidates at large vocabulary sizes. Even with this adjustment, TileLang-TopK remains limited to $V \leq 2^{17}$ due to shared-memory capacity constraints.

\begin{wrapfigure}{r}{0.5\linewidth}
  \centering
  \includegraphics[width=\linewidth, trim=0 0 0 1.5cm, clip]{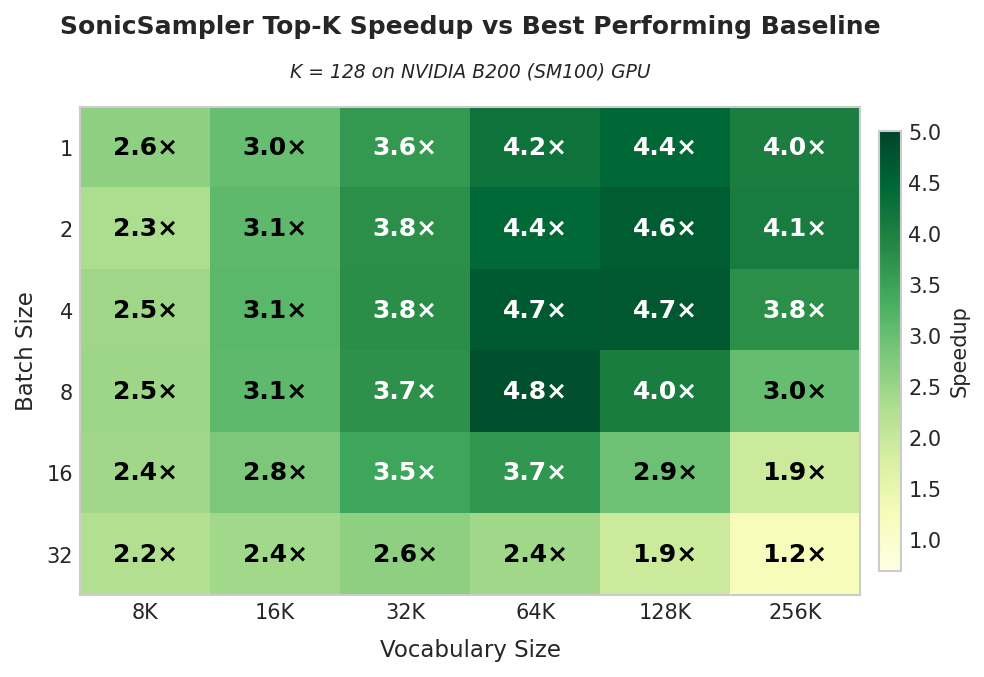}
  \captionof{figure}{\small Top-$k$=128 speedup over the best baseline.}
  \label{fig:topk-speedup}
  \vspace{-1.5em}
\end{wrapfigure}

\paragraph{Triton-TopK.}
The original implementation uses a fused value-index packing scheme that restricts the maximum supported vocabulary size to $V \leq 2^{15}$. We retain this implementation as-is and report configurations beyond this limit as unsupported.

As shown in \Cref{fig:topk-latency}, SonicSampler achieves the lowest latency across all evaluated configurations and scales robustly with both vocabulary size and batch size. In contrast, existing approaches either incur higher constant overhead or degrade more sharply as vocabulary size increases. The advantage is most pronounced at moderate batch sizes and larger vocabularies, where global reduction increasingly dominates runtime. At higher batch sizes, where the regimes become increasingly compute-bound, the relative speedup narrows but remains substantial.


\paragraph{Speedup.}
\Cref{fig:topk-speedup} summarizes the relative gains, showing that SonicSampler consistently outperforms the best across baselines by $2$--$5\times$ in most configurations. Detailed comparisons against FlashInfer and TileLang are deferred to \Cref{app:topk}. These results demonstrate that our two-stage Top-$k$ algorithm provides both lower latency and more robust scaling, forming the foundation of SonicSampler's end-to-end performance gains.

\begin{table}[H]
    \centering
    \small
    \vspace{6pt}

    \begin{tabular}{@{}r
      *{2}{r}@{\hskip 10pt}
      *{2}{r}@{\hskip 10pt}
      *{2}{r}@{\hskip 10pt}
      *{2}{r}@{\hskip 10pt}
      *{2}{r}@{\hskip 10pt}
      *{2}{r}@{}}
    \toprule
    & \multicolumn{2}{c}{$V\!=\!8\text{K}$}
    & \multicolumn{2}{c}{$V\!=\!16\text{K}$}
    & \multicolumn{2}{c}{$V\!=\!32\text{K}$}
    & \multicolumn{2}{c}{$V\!=\!64\text{K}$}
    & \multicolumn{2}{c}{$V\!=\!128\text{K}$}
    & \multicolumn{2}{c}{$V\!=\!256\text{K}$} \\
    \cmidrule(lr){2-3}\cmidrule(lr){4-5}\cmidrule(lr){6-7}
    \cmidrule(lr){8-9}\cmidrule(lr){10-11}\cmidrule(lr){12-13}
    $B$ & bit & adp & bit & adp & bit & adp & bit & adp & bit & adp & bit & adp \\
    \midrule
     1 & \textbf{1.00} & 1.33 & \textbf{1.00} & 1.16 & \textbf{1.00} & 1.20 & \textbf{1.00} & 1.24 & \textbf{1.00} & 1.22 & \textbf{1.00} & 1.14 \\
     2 & \textbf{1.13} & 1.37 & \textbf{1.02} & 1.21 & \textbf{1.00} & 1.20 & \textbf{1.00} & 1.24 & \textbf{1.00} & 1.23 & \textbf{1.01} & 1.14 \\
     4 & \textbf{1.15} & 1.37 & \textbf{1.02} & 1.23 & \textbf{1.00} & 1.20 & \textbf{1.01} & 1.25 & \textbf{1.01} & 1.23 & \textbf{1.14} & 1.15 \\
     8 & \textbf{1.15} & 1.37 & \textbf{1.04} & 1.23 & \textbf{1.01} & 1.20 & \textbf{1.02} & 1.26 & 1.26 & \textbf{1.23} & 1.57 & \textbf{1.47} \\
    16 & \textbf{1.17} & 1.37 & \textbf{1.14} & 1.25 & \textbf{1.11} & 1.21 & 1.41 & \textbf{1.36} & 1.87 & \textbf{1.72} & \textbf{2.33} & 2.50 \\
    32 & \textbf{1.28} & 1.39 & \textbf{1.33} & 1.46 & 1.55 & \textbf{1.47} & \textbf{2.13} & 2.23 & \textbf{2.71} & 3.07 & \textbf{3.83} & 4.42 \\
    \bottomrule
    \end{tabular}

    \vspace{4pt}
    \caption{%
      Normalized latency of \textsc{bitonic} (bit) vs.\ \textsc{adaptive} (adp) sorting strategies
      across vocab sizes and batch sizes on B200 ($k=128$).
      All values are divided by $\min(\text{bitonic}_1,\,\text{adaptive}_1)$ for
      the corresponding architecture and vocab size, so 1.00 anchors the fastest
      \texttt{batch\_size\,=\,1} configuration.
      \textbf{Bold} marks the lower (better) value per cell; ties bold both.%
    }
    \label{tab:normalized-latency-b200}
\end{table}

\paragraph{Strategy Selection.}
Our top-$k$ kernel supports two strategies, namely an optimized bitonic variant and an adaptive variant.
As shown in \Cref{tab:normalized-latency-b200}, neither strategy dominates across all batch sizes.
Their relative performance varies with different batch sizes. SonicSampler benchmarks both strategies offline and selects the faster variant for each batch size at runtime. We also show a similar trend holding for H100 in \Cref{tab:normalized-latency-h100}.

\subsection{Effective Losslessness under Bounded Top-$k$}
\label{exp:correctness}

SonicSampler is exact for the full logit processing pipeline, including temperature scaling, top-$k$ selection, frequency/presence penalties, etc. The only potential approximation arises in nucleus sampling when the post-top-$p$ support exceeds 128 tokens, as our kernel assumes at most 128 nonzero entries.

\begin{figure}[h]
    \centering
    \begin{subfigure}[t]{0.765\textwidth}
        \centering
        \includegraphics[width=\linewidth]{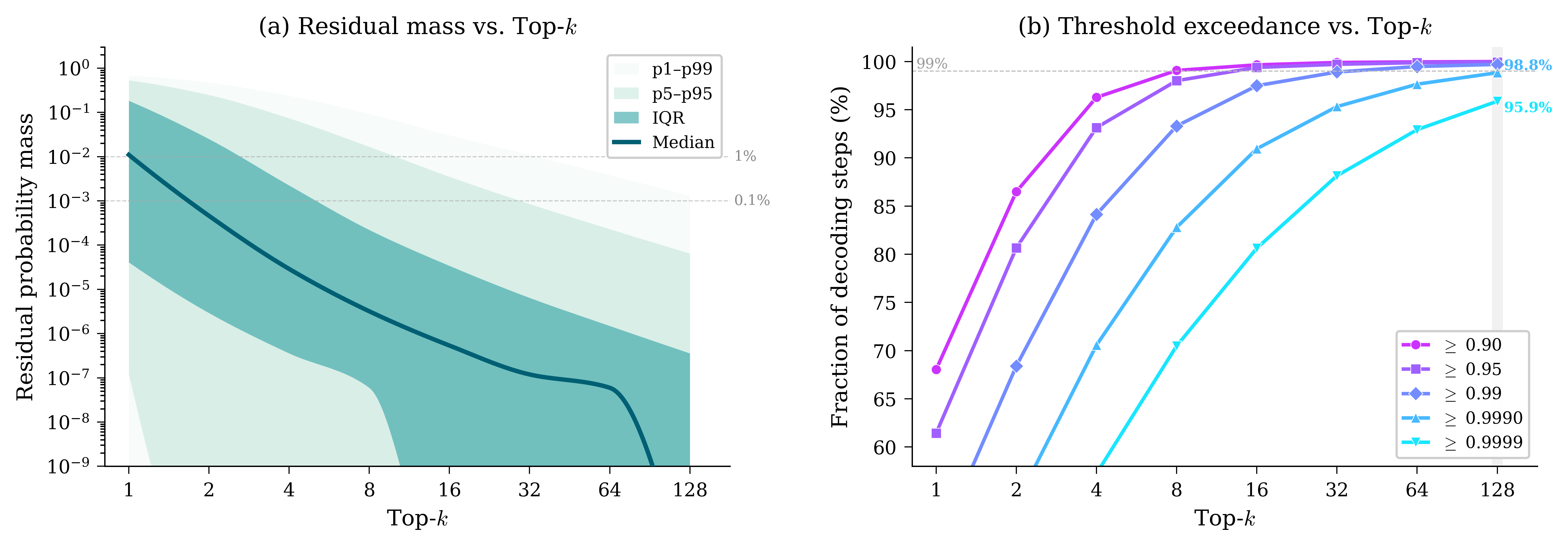}
    \end{subfigure}
    \hfill
    \begin{subfigure}[t]{0.225\textwidth}
        \centering
        \includegraphics[width=\linewidth]{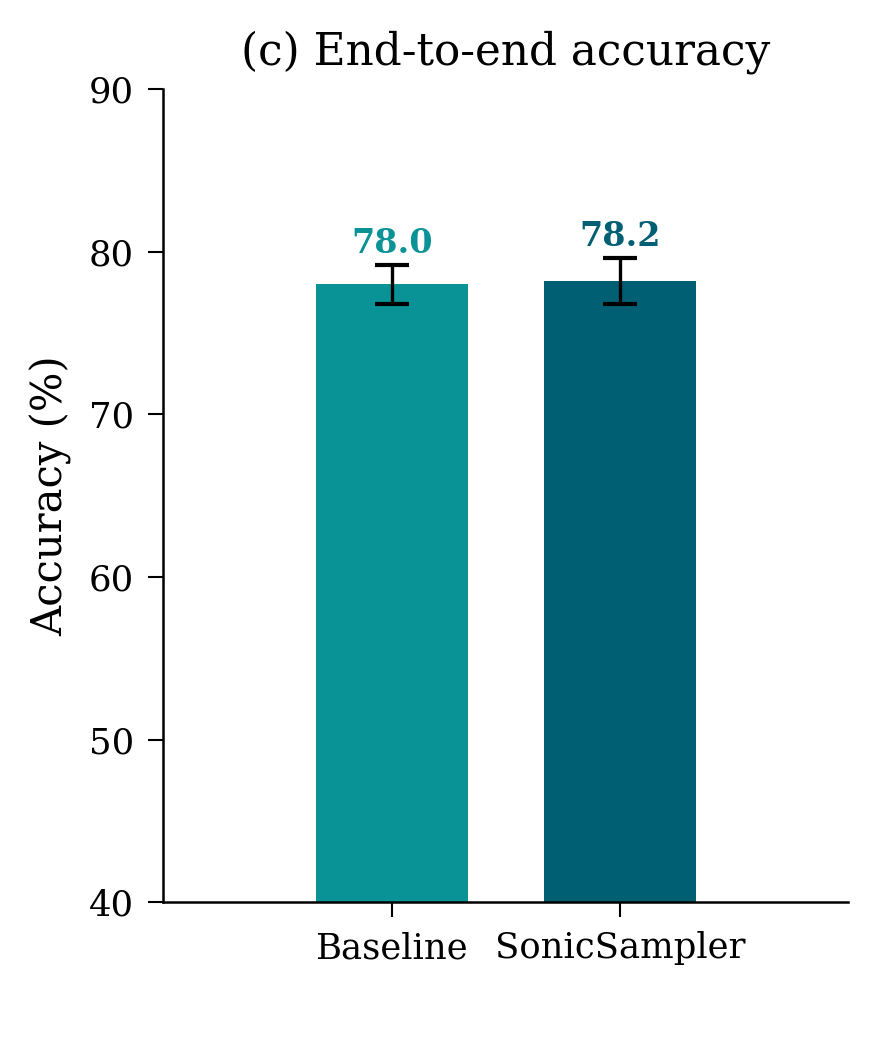}
    \end{subfigure}
    \caption{\small
        Effective losslessness and end-to-end validation.
        (a) Residual probability mass outside top-$k$ on DeepSeek v3.1 logits.
        (b) Fraction of steps where top-$k$ mass exceeds top-$p$ thresholds;
        (c) End-to-end accuracy of DeepSeek v3.1 on GPQA-Diamond.
    }
    \label{fig:quality}
\end{figure}

\paragraph{Support truncation analysis.}
We quantify this effect on DeepSeek v3.1 \citep{deepseekai2025deepseekv3technicalreport} logits. As shown in \Cref{fig:quality} (a), the residual probability mass outside top-$k$ decays rapidly, reaching $\sim10^{-8}$ at $k=128$. \Cref{fig:quality} (b) shows that at $k=128$, over 98.7\% of decoding steps already exceed $p \ge 0.999$, and 96.1\% exceed $p \ge 0.9999$, while standard thresholds (e.g., $p=0.9, 0.95$) are effectively always satisfied.
Thus, although top-$p$ under bounded top-$k$ is theoretically lossy, the excluded mass is negligible in practice, making our kernels \textit{effectively lossless}. Results on other models can be found in \Cref{app:mass_validation}.

\paragraph{End-to-end accuracy.}
To further validate that this has no practical impact, we evaluate end-to-end task performance. As shown in \Cref{fig:quality} (c), on GPQA-Diamond with DeepSeek v3.1 (temperature 0.6, top-$p$ 0.9), SonicSampler matches the Torch-based baseline. 
The negligible difference, well within statistical variance, confirms that the bounded-support assumption does not affect downstream accuracy, reinforcing that SonicSampler is effectively lossless in real workloads.
\Cref{app:acc} presents results on other datasets.

\section{Conclusion and Future Directions}

We presented SonicSampler, a CUDA Graph-compatible framework that fuses the full sampling pipeline into a single workload-aware kernel, achieving the lowest kernel latency and consistent end-to-end speedups across diverse settings. 
These results show that optimizing sampling directly yields system-level gains and requires co-design across operator efficiency, vertical fusion, and support for heterogeneous workloads. 
Future work includes extending to beam search and deeper integration with upstream and downstream components.

\bibliographystyle{plainnat}
\bibliography{references}

\newpage

\appendix
\crefalias{section}{appendix}
\crefalias{subsection}{appendix}

\Crefname{appendix}{Appendix}{Appendices}
\crefname{appendix}{appendix}{appendices}

\section{Extended Related Work}
\label{app:related}

We provide a more detailed survey of prior work on LLM sampling, top-$k$ selection, and kernel fusion for completeness.

\paragraph{FlashInfer.} FlashInfer~\citep{flashinfer2024} provides GPU kernels for top-$k$ and top-$p$ sampling using a sorting-free, pivot-based formulation. Kernels are organized around specific operations (e.g., \texttt{min\_p\_sampling\_from\_probs}), with separate code paths for greedy and stochastic sampling. The pivot selection procedure is data-dependent, with performance characteristics that vary across temperature settings, and each iteration performs block-level reduction over the full vocabulary.

\paragraph{XGrammar.} XGrammar~\citep{dong2025xgrammarflexibleefficientstructured} enables efficient grammar-constrained decoding via byte-level pushdown automaton masking, with CPU-side mask generation overlapped with GPU inference to hide latency. The grammar mask is computed on the CPU and transferred to the GPU, where it is applied to logits as a standalone kernel prior to sampling.

\paragraph{Mojo / MAX.} \cite{modular2025max} provides a fused GPU kernel (\texttt{topk\_fused\_sampling}) that jointly handles top-$k$ filtering, nucleus sampling, temperature scaling, and stochastic selection in a single launch. Mixed greedy/stochastic batches are routed entirely to the stochastic path when any non-greedy request is present. The fused kernel does not return intermediate probability distributions, which makes integration with speculative verification tricky. MAX's graph compiler supports operator fusion via MLIR-based IR, with the sampling stage executing as a separate compiled graph from the model forward pass.

\paragraph{MegaKernel.} The MegaKernel approach~\citep{cheng2025miragepersistentkernelcompiler} pursues end-to-end kernel fusion by compiling multi-layer LLM inference into a single megakernel, thereby reducing launch overhead and enabling cross-operator pipelining. While this is aligned with SonicSampler's broader fusion perspective, MegaKernel focuses on the model forward pass rather than the sampling stage, and thus does not directly target challenges such as large-vocabulary top-$k$ reduction or the composition of heterogeneous sampling strategies (e.g., greedy, nucleus, and grammar-constrained sampling) in dynamically batched settings.

\paragraph{Top-$k$ Selection Algorithms.} Efficient top-$k$ selection over modern LLM vocabularies ($V \geq 2^{17}$) remains an open challenge, with existing approaches trading off between CUDA Graph compatibility, data type flexibility, and scalability:

\textit{(1) RadiK.}
RadiK~\citep{radik2024} implements multi-pass radix selection, where each radix pass is delimited by CPU-GPU synchronization, rendering it incompatible with CUDA Graph capture. Its public release additionally supports only \texttt{fp32}, limiting its applicability in mixed-precision inference pipelines prevalent in modern deployments.

\textit{(2) Triton Streaming Bitonic.}
Triton's streaming bitonic top-$k$, while being the most proximal to our implementation in that we share the underlying bitonic network implementation, is currently constrained to $V < 2^{15}$ due to its fused value-index packing scheme, which reserves 16 bits for indices within a wider sorting key. Additionally, the block-wise streaming design forces it to loop over vocabulary blocks within a single program, which limits opportunities for cross-block parallelism.

\textit{(3) FlashInfer Top-$k$ Variants.} 
FlashInfer provides two top-$k$ paths selected by a heuristic wrapper that considers $k$, vocabulary size, and data type. \texttt{FilteredTopK} uses a single-CTA, shared-memory-intensive 8-bit coarse histogram with iterative radix refinement, but is constrained to $k \leq 2048$ with 128KB shared memory, and is further limited to vocabulary sizes below dtype-dependent thresholds (16K for \texttt{fp16}/\texttt{bf16}, 32K for \texttt{fp32}). Beyond these thresholds, the wrapper falls back to \texttt{Multi-CTA RadixTopK}, which chunks rows across thread blocks with global histogram coordination, with associated atomic and synchronization costs.

\textit{(4) TileLang.} TileLang~\citep{wang2025tilelangcomposabletiledprogramming} provides a tile-aware multi-pass radix selection implementation that proves competitive with our approach (\Cref{sec:experiments}). Its top-$k$ kernel assigns a single thread block per batch element, iterating over vocabulary tiles sequentially within each block. In contrast, our two-stage decomposition distributes vocabulary blocks across independent thread blocks in the first phase, increasing SM occupancy, followed by a lightweight merge over partial results.

\textit{(5) Qrita.} 
Qrita~\citep{park2026qritahighperformancetopktopp} combines Gaussian $\sigma$-truncation, which prunes the search space by thresholding logits to statistical outliers, with a quaternary pivot search that halves iteration count relative to binary search. However, the single-CTA Triton implementation limits SM utilization for large vocabularies, as each program must sequentially iterate over vocabulary blocks exceeding the thread block size. Performance is also data-dependent, with $\sigma$-truncation effectiveness conditioned on the logit distribution conforming to its Gaussian assumption. As a standalone top-$k$ primitive, Qrita does not address fusion with the surrounding logit processing pipeline.

\paragraph{Large-Scale Selection.} The need for fast top-$k$ over large $N$ extends beyond vocabulary selection. DeepSeek-V3.2's Lightning Indexer~\citep{deepseekai2025deepseekv32pushingfrontieropen} performs sparse attention token selection over context lengths approaching $10^5$, driving demand for efficient top-$k$ implementations in frameworks such as TileLang and FlashInfer. This underscores the broader applicability of scalable top-$k$ algorithms. Yet within the sampling context, top-$k$ is uniquely positioned: it is preceded by a dense chain of element-wise logit transformations and followed by probability normalization and stochastic selection. This structure presents an opportunity for vertical fusion that standalone top-$k$ implementations, designed as isolated primitives, cannot exploit. 

\paragraph{FlashSampling.} FlashSampling~\citep{ruiz2026flashsampling} accelerates sampling by fusing the LM head with downstream sampling operations, reducing memory traffic between logit generation and token selection. FlashSampling operates at the boundary between the model forward pass and token selection, whereas the post-logit pipeline—logit processing, top-$k$/top-$p$ filtering, and speculative verification—constitutes a separate stage of the inference stack.

\begin{table}[t]
  \centering
  \setlength{\tabcolsep}{5pt}
  \begin{tabular}{@{}lccccccc@{}}
  \toprule
  & Grammar & Penalties & Bias & Truncation & Verify & CG & Mixed G/S \\
  \midrule
  FlashInfer            & $\times$     & $\times$     & $\times$     & $\checkmark$ & $\checkmark$ & $\times$
& $\times$     \\
  Torch                 & $\checkmark$ & $\checkmark$ & $\checkmark$ & $\checkmark$ & $\checkmark$ & $\checkmark$
& $\times$     \\
  Mojo                  & $\times$     & $\times$     & $\times$     & $\checkmark$ & $\times$     & $\times$
& $\times$     \\
  \midrule
  \textbf{SonicSampler} & \checkmark   & \checkmark   & \checkmark   & \checkmark   & \checkmark   & \checkmark
& \checkmark   \\
  \bottomrule
  \end{tabular}
  \caption{
  Feature comparison of sampling kernels. $\checkmark$ indicates full support, $\times$ indicates no support.
  `Penalties' denotes repetition, frequency, and presence penalties; `Truncation' denotes top-$k$, top-$p$, and
min-$p$ probability truncation;
  `Verify' denotes speculative verification; `CG' denotes CUDA Graph compatibility; and `Mixed G/S' denotes mixed
greedy/stochastic batch support.
  }
  \label{tab:feature_comparison}
\end{table}

\section{Feature Comparison}
\label{app:feature_comparison}

\Cref{tab:feature_comparison} summarizes the feature coverage across existing implementations. 
SonicSampler is the only implementation providing unified coverage of the complete sampling pipeline while maintaining CUDA Graph compatibility and supporting mixed greedy/stochastic batches.

\section{Background: Detailed Formulation}
\label{app:background}

We formalize the logit processing, sampling, and verification operations that constitute the LLM sampling pipeline. Let $\mathbf{x} \in \mathbb{R}^V$ denote raw logits over a vocabulary of size $V$, and $\mathbf{p} = \text{Softmax}(\mathbf{x})$ the corresponding probability distribution.

\subsection{Logit Processing Operations}

\paragraph{Grammar Bit-Masking.} Constrained decoding enforces structural validity by masking logits corresponding to tokens that would violate grammatical constraints. Given a binary mask $\mathbf{m} \in \{0, 1\}^V$ derived from a finite-state machine state, we apply:
\begin{equation}
    x_i \leftarrow \begin{cases} x_i & \text{if } m_i = 1 \\ -\infty & \text{otherwise} \end{cases}
\end{equation}

\paragraph{Repetition Penalties.}
To discourage degenerate repetition in open-ended generation, three complementary penalties are applied to previously generated tokens.
Let $c_i \in \mathbb{N}_{0}$ denote the number of times token~$i$ has appeared in the decoded sequence so far, and let $\alpha_{\mathrm{rep}} \ge 1$, $\beta_{\mathrm{freq}} \ge 0$, and $\gamma_{\mathrm{pres}} \ge 0$ be the repetition, frequency, and presence penalty coefficients, respectively.

The \emph{multiplicative repetition penalty} contracts positive logits and amplifies negative logits for any previously generated token, preserving the sign-dependent asymmetry needed to maintain the correct probability ordering:
\begin{equation}
    \tilde{x}_i =
    \begin{cases}
        x_i \,/\, \alpha_{\mathrm{rep}}  & \text{if } c_i > 0 \;\text{and}\; x_i > 0, \\
        x_i \cdot \alpha_{\mathrm{rep}}   & \text{if } c_i > 0 \;\text{and}\; x_i \le 0, \\
        x_i                               & \text{otherwise}.
    \end{cases}
\end{equation}
The \emph{frequency penalty} then applies an additive reduction proportional to each token's occurrence count, and the \emph{presence
penalty} applies a flat additive reduction to every token that has appeared at least once, yielding the final penalized logit:
\begin{equation}
    x_i \;\leftarrow\; \tilde{x}_i
        \;-\; \beta_{\mathrm{freq}}\cdot\, c_i
        \;-\; \gamma_{\mathrm{pres}}\cdot\, \mathbb{I}[\,c_i > 0\,].
\end{equation}
Setting $\alpha_{\mathrm{rep}} = 1$, $\beta_{\mathrm{freq}} = 0$,
$\gamma_{\mathrm{pres}} = 0$ recovers the unpenalized logits.

\paragraph{Logit Bias.} Request-specific additive biases $\mathbf{b} \in \mathbb{R}^V$: 
\begin{equation}
    x_i \leftarrow x_i + b_i
\end{equation}

\paragraph{Temperature.} Temperature $\tau > 0$ modulates distribution entropy:
\begin{equation}
    x_i \leftarrow x_i / \tau
\end{equation}
Lower temperatures sharpen the distribution toward the mode; higher temperatures increase diversity.

\subsection{Probability Truncation}

\paragraph{Top-$k$.} Restricts the candidate set to the $k$ highest-probability tokens, which is analogous to selecting the $k$ highest valued logits, as Softmax is inherently a monotonic and order-preserving transformation:
\begin{equation}
    \mathcal{T}_k = \{i : \text{rank}(x_i) \leq k\}
\end{equation}

\paragraph{Top-$p$ (Nucleus).} Select the minimal set whose cumulative probability exceeds the threshold $p$~\citep{holtzman2020curious}:
\begin{equation}
    \mathcal{T}_p = \arg\min_{S} |S| \quad \text{s.t.} \quad \sum_{i \in S} p_i \geq p, \; p_i \geq p_j \; \forall i \in S, j \notin S
\end{equation}

\paragraph{Min-$p$.} \label{para:min_p}
Establishes a probability floor relative to the maximum. The naive formulation requires computing the full softmax:
\begin{equation}
    \mathcal{T}_{\theta} = \{i : p_i \geq \theta \cdot \max_j p_j\}
\end{equation}
However, we observe that this criterion admits a simplification in logit space. Since $\ln p_i = x_i - \ln \sum_j e^{x_j}$, we have:
\begin{align}
    p_i \geq \theta \cdot \max_j p_j \quad &\Leftrightarrow \quad \ln p_i \geq \ln \theta + \max_j \ln p_j \nonumber \\
    &\Leftrightarrow \quad x_i \geq M + \ln \theta
    \label{eq:min_p_trick}
\end{align}
where $M = \max_j x_j$. This reduces min-$p$ filtering to a simple threshold comparison against the maximum logit, avoiding explicit softmax computation until after truncation. In fact, we leverage this critical simplification towards optimizing the min-$p$ processing within SonicSampler.

\subsection{Sampling}
\label{app:sampling}

\paragraph{Gumbel-Max Reparameterization.} Sampling from a categorical distribution can be reformulated as a deterministic $\arg\max$ over perturbed logits. If $\mathbf{g} = (g_1, \ldots, g_V)$ are i.i.d.\ samples from $\text{Gumbel}(0, 1)$, then:
\begin{equation}
    x^* = \arg\max_i (g_i + \ln p_i) \sim \text{Categorical}(\mathbf{p})
\end{equation}
Since $g_i = -\ln z_i$ where $z_i \sim \text{Exp}(1)$, and noting that the log-partition function is constant across indices, we obtain:
\begin{equation}
    x^* = \arg\max_i (x_i - \ln z_i)
\end{equation}
This formulation unifies stochastic sampling with the $\arg\max$ operation used in greedy decoding, enabling a single code path where greedy selection corresponds to $z_i = 1$ for all $i$.

\subsection{Speculative Verification}

In speculative decoding, a draft model proposes a sequence of $L$ candidate tokens $(d_1, \ldots, d_L)$ with probabilities $q(d_\ell)$. The target model computes probabilities $p(d_\ell)$ for each candidate. Verification proceeds sequentially: token $d_\ell$ is accepted if~\citep{leviathan2023fast}:
\begin{equation}
    u_\ell \leq \frac{p(d_\ell)}{q(d_\ell)}, \quad u_\ell \sim \text{Uniform}(0, 1)
\end{equation}
Upon rejection at position $\ell$, a correction token is sampled from the residual distribution:
\begin{equation}
    \tilde{p}(x) = \text{normalize}\left(\max\{0, p(x) - q(x)\}\right)
\end{equation}
The first rejected position determines the acceptance length, and decoding continues from the corrected token.

\paragraph{Draft-to-Target Mapping.} Tree-structured speculation methods such as EAGLE-3~\citep{li2025eagle} require mapping between draft and target token indices during verification. Let $\pi: \{1, \ldots, L\} \to \{1, \ldots, L'\}$ denote the mapping from draft positions to target positions. This indirection is applied during probability lookup and token writeback in the verification kernel.






\section{Generalized Packing Transform for Alternating Sub-Tiles}
\label{app:alternating_indices}

The basic packing transform $\psi(x, i, n) = \phi(x) \cdot 2^{16} + (n - 1) \oplus i$ encodes a value--index pair into a single \texttt{uint32} for tile-local sorting.
In Stage~2, the scratchpad contains $Z_v' \cdot k$ packed entries across $Z_v'$ vocabulary blocks (where $Z_v' = 2^{\lceil \log_2 Z_v \rceil}$), and the bitonic fold expects alternating ascending--descending sub-tiles.
We therefore assign \emph{relative indices} across the full scratchpad such that the packed representation preserves each block's sort order from Stage~1.

\paragraph{Forward Transform $\Psi$.}
Let $j \in [0, Z_v')$ be the block index, $r \in [0, k)$ the position within block~$j$, $\pi_j = j \bmod 2$ the block parity, and $N = Z_v' \cdot k$ the total scratchpad width.
Recall that Stage~1 writes block~$j$'s candidates in ascending order when $\pi_j = 0$ and descending order when $\pi_j = 1$.
Define the alternating mask
\begin{equation}
    \mu(j) \;=\; (N - 1) \oplus \bigl((1 - \pi_j)(k - 1)\bigr),
    \label{eq:alternating_mask}
\end{equation}
and the generalized relative index
\begin{equation}
    \Psi(j,\, r) \;=\; (jk + r) \oplus \mu(j).
    \label{eq:psi_general}
\end{equation}
Expanding by parity and writing $\bar{j} = (Z_v' - 1) - j$ for the bit-complement of $j$ within $\lceil \log_2 Z_v' \rceil$ bits:
\begin{equation}
    \Psi(j,\, r) \;=\;
    \begin{cases}
        \bar{j}\, k + r          & \text{if } \pi_j = 0\ \text{(ascending)}, \\[4pt]
        \bar{j}\, k + (k - 1 - r) & \text{if } \pi_j = 1\ \text{(descending)}.
    \end{cases}
    \label{eq:psi_expanded}
\end{equation}
For ascending blocks, $\Psi$ increases with $r$, matching the ascending value order; for descending blocks, $\Psi$ decreases with $r$, matching the descending value order.
The block complement $\bar{j}$ arranges the sub-tiles into the interleaved pattern expected by the bitonic fold network.

Stage~2 replaces each entry's tile-local packed index with $\Psi(j, r)$ while separately pre-computing the absolute vocabulary indices
\begin{equation}
    \text{abs}[j,\, r] \;=\; j \cdot B_N + \bigl((p_{j,r}\ \land \texttt{0xFFFF}) \oplus (B_N - 1)\bigr),
    \label{eq:absolute_index}
\end{equation}
where $p_{j,r}$ is the original packed \texttt{uint32} from Stage~1 and the XOR with $(B_N - 1)$ inverts the tile-local index complement from the basic packing transform $\psi$.

\paragraph{Inverse Transform $\Psi^{-1}$.}
After the bitonic reduction selects $k$ winners with relative indices $\{\ell'_0, \ldots, \ell'_{k-1}\}$, the original block and position are recovered as follows.
First, apply the total complement:
\begin{equation}
    \tilde{\ell} \;=\; \ell' \oplus (N - 1).
    \label{eq:total_complement}
\end{equation}
Then extract the block index and correct the within-block position:
\begin{align}
    \hat{j} &\;=\; \tilde{\ell} \gg \kappa, \label{eq:extract_block} \\[4pt]
    \hat{r} &\;=\; \bigl(\tilde{\ell}\ \land (k - 1)\bigr) \oplus \bigl((1 - \hat{j} \bmod 2) \cdot (k - 1)\bigr), \label{eq:correct_position}
\end{align}
where $\kappa = \log_2 k$.
The position correction in \cref{eq:correct_position} re-applies the parity-dependent index flip from $\mu(j)$, yielding the original within-block position $r$ for both parities.
The absolute vocabulary index is then recovered via a gather:
\begin{equation}
    a \;=\; \text{abs}[\hat{j},\, \hat{r}] \;=\; \text{abs}\bigl[\hat{j}\, k + \hat{r}\bigr].
    \label{eq:gather_absolute}
\end{equation}

\paragraph{Correctness.}
We verify the round-trip $\Psi^{-1} \circ \Psi = \text{id}$ for both parities.
For an ascending block ($\pi_j = 0$):
\begin{align}
    \ell' &= \bar{j}k + r \\[4pt]
    \tilde{\ell} &= jk + (k{-}1{-}r)
\end{align}
which yields $\hat{j} = j$ and:
\begin{equation}
    \hat{r} = (k{-}1{-}r) \oplus (k{-}1) = r
\end{equation}

For a descending block ($\pi_j = 1$):
\begin{align}
    \ell' &= \bar{j}k + (k{-}1{-}r) \\[4pt]
    \tilde{\ell} &= jk + r
\end{align}

which gives $\hat{j} = j$ and $\hat{r} = r \oplus 0 = r$. In both cases, the original $(j, r)$ pair is recovered exactly.

\section{Supplementary Algorithms}
\label{app:pseudocode}

This appendix provides the complete pseudocode for the algorithms referenced but not fully detailed in the main text.


\begin{algorithm}[H]
\caption{Hypercube Merge}\label{alg:hypercube_merge}
\small
\begin{algorithmic}[1]
\Require Hypercube $\mathbf{v} \in \mathbb{N}^{2^n}$, pair $\rho$, flip $f$, stage $s$, total $n$
\Ensure Merged $\mathbf{v}' \in \mathbb{N}^{2^n}$
\For{$i \gets 0$ to $s - 1$}
    \State $\mathbf{v} \gets \textsc{CAS}(\mathbf{v}, \rho, f,\, s - i - 1,\, n)$
\EndFor
\State \Return $\mathbf{v}$
\end{algorithmic}
\end{algorithm}

\begin{algorithm}[H]
\caption{Bitonic Reduction (Cross-Tile Merge)}\label{alg:bitonic_reduction}
\small
\begin{algorithmic}[1]
\Require Block $\mathbf{B} \in \mathbb{N}^{R \times k}$, rows $R$, target $k$, total columns $N$
\Ensure Reduced candidates $\mathbf{y} \in \mathbb{N}^k$
\State $n \gets \log_2 N + 1$,\quad $\kappa \gets \log_2 k$,\quad $F \gets \log_2 R$
\State $\rho \gets \{0,1\}$
\State $\mathbf{y} \gets \textsc{BitonicFold}(\mathbf{B},\, \rho,\, R,\, k,\, F,\, \delta{=}1,\, n,\, \kappa)$
\State \Return $\mathbf{y}$
\end{algorithmic}
\end{algorithm}


\begin{algorithm}[H]
\caption{Bitonic Selection (Stage 1 Write-Out)}\label{alg:bitonic_selection}
\small
\begin{algorithmic}[1]
\Require Scratchpad $Y$, batch $b$, block $j$, packed $\mathbf{p} \in \mathbb{N}^{B_N}$, target $k$
\Ensure $k$ candidates written to $Y[b, j \cdot k : (j{+}1)k]$
\State $\mathbf{y} \gets \textsc{BitonicTopK}(\mathbf{p},\, j \bmod 2,\, k,\, B_N)$
    \Comment{Alternating order}
\State $\textsc{Store}(Y + b \cdot S_Y + j \cdot k,\, \mathbf{y})$
\end{algorithmic}
\end{algorithm}

  \begin{algorithm}[H]
  \caption{Bitonic Top-$k$ Selection}\label{alg:bitonic_topk}
  \small
  \begin{algorithmic}[1]
  \Require Packed values $\mathbf{p} \in \mathbb{N}^{B_N}$, order $\delta \in \{0,1\}$, target $k$
  \Ensure Top-$k$ candidates $\mathbf{y} \in \mathbb{N}^k$
  \State $n \gets \log_2 B_N$,\quad $\kappa \gets \log_2 k$
  \State $F \gets n - \kappa - 1$,\quad $R \gets 2^F$
  \State $\mathbf{H} \gets \textsc{Reshape}(\mathbf{p},\ [2]^n)$ \Comment{Hypercube view}
  \State $\rho \gets \{0,1\}$
  \For{$s \gets 1$ to $\kappa$} \Comment{Sort $k$-width slices}
      \State $f \gets \textsc{Expand}(\rho,\ n{-}s{-}1,\ s)$
      \State $\mathbf{H} \gets \textsc{HypercubeMerge}(\mathbf{H}, \rho, f, s, n)$
  \EndFor
  \State $\mathbf{H} \gets \max(\mathbf{H},\, \text{axis}=F)$ \Comment{Reduce folds}
  \If{$F > 0$}
      \State $f \gets \textsc{Expand}(\rho,\ F{-}1,\ \kappa)$
      \State $\mathbf{H} \gets \textsc{HypercubeMerge}(\mathbf{H}, \rho, f, \kappa, n{-}1)$
      \State $\mathbf{G} \gets \textsc{Reshape}(\mathbf{H},\ [R,\, k])$
      \State $\mathbf{y} \gets \textsc{BitonicFold}(\mathbf{G}, \rho, R, k, F, \delta)$
  \Else
      \State $\mathbf{H} \gets \textsc{HypercubeMerge}(\mathbf{H}, \rho, \delta, \kappa, n{-}1)$
      \State $\mathbf{y} \gets \textsc{Reshape}(\mathbf{H},\ [k])$
  \EndIf
  \State \Return $\mathbf{y}$
  \end{algorithmic}
  \end{algorithm}


\begin{algorithm}[H]
\caption{Radix Selection (Stage 1 Write-Out)}\label{alg:radix_selection}
\small
\begin{algorithmic}[1]
\Require Scratchpad $Y$, batch $b$, block $j$, encoded $\mathbf{e}$, threshold $\theta$, margin $m$, indices $\mathbf{I}$, target $k$, tile size $B_N$
\Ensure $k$ candidates written to $Y[b, j \cdot k : (j{+}1)k]$
\State $\mathbf{p} \gets \psi(\mathbf{e},\, \mathbf{I},\, B_N)$ \Comment{Pack with inverted indices}
\State $\mathbf{q} \gets (\mathbf{e} = \theta) \mathbin{\&} (\textsc{CumSum}(\mathbf{e} = \theta) \leq m)$
    \Comment{Margin tie-breaking}
\State $\mathbf{M} \gets (\mathbf{e} > \theta) \mathbin{|} \mathbf{q}$ \Comment{Selection mask}
\State $\text{pos} \gets \textsc{CumSum}(\mathbf{M}) - \mathbf{M}$
    \Comment{Coalesced positions}
\State $\textsc{MaskedStore}(Y + b \cdot S_Y + j \cdot k + \text{pos},\, \mathbf{p},\, \text{mask}{=}\mathbf{M})$
\end{algorithmic}
\end{algorithm}


\begin{algorithm}[H]
\caption{Sparse Threshold (Two-Pass Radix)}\label{alg:sparse_threshold}
\small
\begin{algorithmic}[1]
\Require Encoded $\mathbf{e} \in \mathbb{N}^{B_N}$, mask $\mathbf{M}$, initial threshold $\theta_0$, target $k$
\Ensure Threshold $\theta$, margin $m$
\State $\beta \gets 5$,\quad $\mathcal{B} \gets 32$,\quad $\mu \gets \mathcal{B} - 1$
    \Comment{5-bit radix with 32 bins}
\State \Comment{\textit{Pass 1: Mid-span bits [5:10)}}
\State $\mathbf{v} \gets ((\mathbf{e} \gg \beta) \mathbin{\&} \mu) \oplus \mu$
    \Comment{Invert for descending order}
\State $h, m \gets \textsc{RadixPartition}(\mathbf{v},\, \mathbf{M},\, k,\, \mathcal{B})$
\State $\theta \gets ((\theta_0 \ll \beta) \mathbin{|} (h \oplus \mu)) \ll \beta$
\If{$m > 0$} \Comment{Pass 2: Low-span bits [0:5)}
    \State $\mathbf{M} \gets \mathbf{M} \mathbin{\&} (\mathbf{v} = h)$
    \State $\mathbf{v} \gets (\mathbf{e} \mathbin{\&} \mu) \oplus \mu$
    \State $t, m \gets \textsc{RadixPartition}(\mathbf{v},\, \mathbf{M},\, m,\, \mathcal{B})$
    \If{$h = 0$ \textbf{and} $t = 0$} \Comment{Degenerate boundary case}
        \State $\theta \gets \theta \mathbin{|} (\mu - 1)$;\quad $m \gets 0$
    \Else
        \State $\theta \gets \theta \mathbin{|} (t \oplus \mu)$
    \EndIf
\Else
    \State $\theta \gets \theta \mathbin{|} \mu$
\EndIf
\State \Return $\theta,\, m$
\end{algorithmic}
\end{algorithm}


\begin{algorithm}[H]
\caption{Stage 1: Tile-Local Reduction Kernel}\label{alg:stage1}
\small
\begin{algorithmic}[1]
\Require Logits $X \in \mathbb{R}^{L \times V}$, indicators $\mathbf{z} \in \mathbb{N}^B$, scratchpad $Y$, tile size $B_N$, target $k$, mode $\in \{\texttt{bitonic}, \texttt{adaptive}\}$
\Ensure Packed top-$k$ candidates per tile in $Y$
\State $b \gets \textsc{ProgramId}(0)$,\quad $j \gets \textsc{ProgramId}(1)$
\State $\mathbf{x} \gets X[b,\, j \cdot B_N : (j{+}1) B_N]$ \Comment{Load logit tile (masked at boundary)}
\State \Comment{\textit{--- Logit Processing Prologue ---}}
\State $\mathbf{x} \gets \textsc{GrammarMask}(\mathbf{x}, z_b)$
\State $\mathbf{x} \gets \textsc{RepetitionPenalty}(\mathbf{x}, z_b)$
\State $\mathbf{x} \gets \textsc{LogitBias}(\mathbf{x}, z_b)$
\State $\mathbf{x} \gets \textsc{Temperature}(\mathbf{x}, z_b)$
\State \Comment{\textit{--- Tile-Local Top-$k$ Reduction ---}}
\If{$z_b$ indicates greedy}
    \State $\textsc{GreedyReduction}(Y, b, j, \mathbf{x})$ \Comment{Scalar $\arg\max$}
\ElsIf{mode $=$ \texttt{adaptive}}
    \State $\textsc{RadixBitonic}(Y, b, j, \mathbf{x}, \mathbf{I}, k, s)$ \Comment{\Cref{alg:radix_bitonic}}
\Else
    \State $\mathbf{p} \gets \psi(\mathbf{x}, \mathbf{I}, B_N)$
    \State $\textsc{BitonicSelection}(Y, b, j, \mathbf{p}, k)$ \Comment{\Cref{alg:bitonic_selection}}
\EndIf
\end{algorithmic}
\end{algorithm}

\begin{algorithm}[H]
\caption{Radix Partition}\label{alg:radix_partition}
\small
\begin{algorithmic}[1]
\Require Values $\mathbf{v}$, mask $\mathbf{M}$, upper bound $u$, bins $\mathcal{B}$
\Ensure Pivot $p$, margin $m$
\State $\mathbf{h} \gets \textsc{CumSum}(\textsc{Histogram}(\mathbf{v},\, \mathcal{B},\, \text{mask}{=}\mathbf{M}))$
\State $\mathbf{b} \gets (\mathbf{h} \leq u)$
\State $p \gets \textsc{Sum}(\mathbf{b})$ \Comment{$\triangleright$ IADD reduction tree}
\State $t \gets \max(\textsc{Select}(\mathbf{b},\, \mathbf{h},\, \mathbf{0}))$ \Comment{$\triangleright$ Masked IMNMX tree}
\State $m \gets \textsc{Select}(p > 0,\, u - t,\, u)$
\State \Return $(p,\, m)$
\end{algorithmic}
\end{algorithm}


\begin{algorithm}[H]
\caption{Stage 2: Cross-Tile Merge and Selection Kernel}\label{alg:stage2}
\small
\begin{algorithmic}[1]
\Require Scratchpad $Y \in \mathbb{N}^{L \times Z_v k}$, indicators $\mathbf{z}$, Gumbel noise $\mathbf{g}$, parameters $(k_b, p_b, n_b)$, target $k$, alternating flag
\Ensure Selected token $t \in \mathbb{N}$, log-probability $w \in \mathbb{R}$
\State $b \gets \textsc{ProgramId}(0)$
\If{$z_b$ indicates greedy}
    \State $t \gets \textsc{GreedyMaximum}(Y, b)$ \Comment{$\arg\max$ over $Z_v$ scalar entries}
\Else
    \State \Comment{\textit{--- Cross-Tile Merge ---}}
    \State Load $[Z_v, k]$ packed entries; remap tile-local to absolute indices
    \If{alternating}
        \State $\mathbf{v}, \mathbf{j} \gets \textsc{BitonicReduction}(Y_b,\, Z_v,\, k)$
            \Comment{\Cref{alg:bitonic_reduction}}
    \Else
        \State $\mathbf{v}, \mathbf{j} \gets \textsc{BitonicTopK}(Y_b,\, 1,\, k,\, Z_v \cdot k)$
    \EndIf
    \State Decode: $\mathbf{v} \gets \phi^{-1}(\mathbf{v}_{\text{upper}})$;\quad $\mathbf{j} \gets \textsc{Gather}(\text{abs.~indices},\, \mathbf{j})$
    \State \Comment{\textit{--- Probability Truncation Epilogue ---}}
    \State $\mathbf{v} \gets \textsc{TopKMask}(\mathbf{v}, k_b)$
    \State $\mathbf{v} \gets \textsc{TopPMask}(\mathbf{v}, p_b)$
        \Comment{Cumulative softmax}
    \State $\mathbf{v} \gets \textsc{MinPMask}(\mathbf{v}, n_b)$
        \Comment{$\log$-domain threshold (\cref{eq:min_p_trick})}
    \State \Comment{\textit{--- Gumbel-Max Token Selection ---}}
    \State $t \gets \mathbf{j}[\arg\max(\mathbf{v} - \mathbf{g}[\mathbf{j}])]$
\EndIf
\State \Return $t$
\end{algorithmic}
\end{algorithm}

\section{Baseline Implementation Details}
\label{app:baseline-modifications}

We evaluate all baselines under matched configurations (precision, input shapes, and hardware).
TileLang uses its native compiler, and RadiK is evaluated as a precompiled CUDA C++ binary.
For a subset of methods, we apply minimal adaptations to ensure correctness and compatibility 
with large vocabulary sizes and \texttt{bfloat16} inputs;
these modifications are conservative and, where measurable, do not degrade and in some cases slightly improve the baseline performance.

\paragraph{FlashInfer.}
FlashInfer's sampling kernel generates Gumbel noise internally from a random seed, which breaks CUDA graph compatibility. 
We modify the kernel to accept precomputed Gumbel noise as an external input, 
enabling CUDA graph capture and ensuring a fair comparison with our approach.
We also adapt the kernel to operate on \texttt{bfloat16} inputs to match our evaluation precision.

\paragraph{TileLang-TopK.}
The original implementation supports only \texttt{float32} inputs. We extend it to support \texttt{bfloat16} by introducing an explicit cast path during input loading and promoting values to \texttt{float32} for intermediate computation.
We additionally increase the shared-memory buffer size from 4,096 to 16,384 entries to avoid silent truncation of candidates at large vocabulary sizes. Even with this adjustment, TileLang-TopK remains limited to $V \leq 2^{17}$ due to shared-memory capacity constraints.

\paragraph{Triton-TopK.}
The original implementation uses a fused value-index packing scheme that restricts the maximum supported vocabulary size to $V \leq 2^{15}$. We retain this implementation as-is and report configurations beyond this limit as unsupported.






\section{Additional Top-$k$ Results}
\label{app:topk}

\begin{figure*}[t]
    \centering
    \includegraphics[width=0.8\linewidth, trim=0 0 0 2cm, clip]{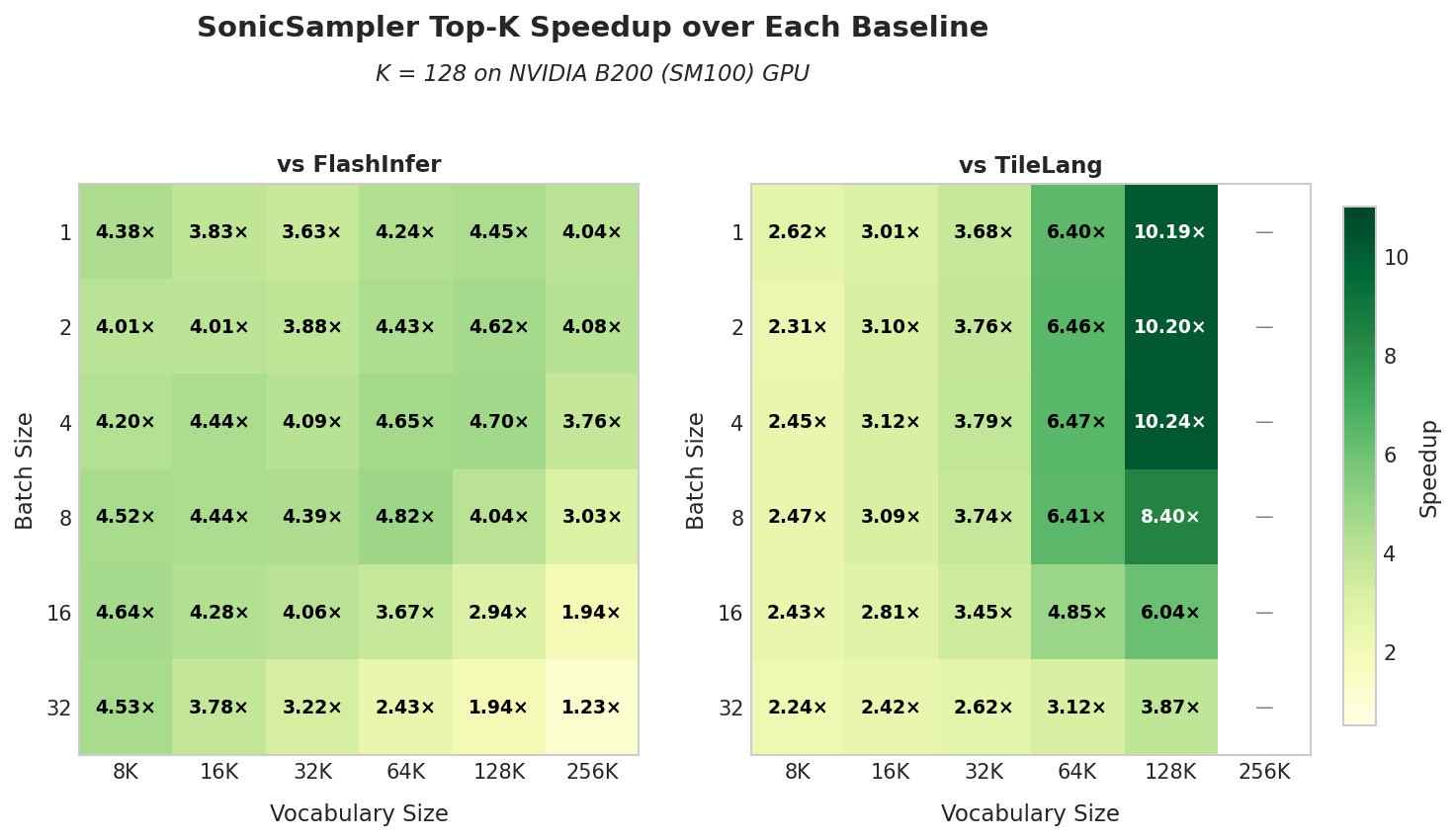}
    \caption{Top-$k$ speedup against FlashInfer and TileLang for $k = 128$ on B200.}
    \label{fig:topk-speedup-fi-tl}
\end{figure*}

\Cref{fig:topk-speedup-fi-tl} reports additional Top-$k$ results comparing SonicSampler against FlashInfer and TileLang.
This figure complements the main-text evaluation by providing a more focused view of the strongest baseline implementations. 
Across the tested settings, SonicSampler consistently outperforms both methods, 
with the relative gains generally increasing as the vocabulary size and workload scale grow.


%

\begin{table}[t]
    \centering
    \label{tab:normalized-latency-h100}
    \small
    
    \vspace{6pt}
    \begin{tabular}{@{}r
      *{2}{r}@{\hskip 10pt}
      *{2}{r}@{\hskip 10pt}
      *{2}{r}@{\hskip 10pt}
      *{2}{r}@{\hskip 10pt}
      *{2}{r}@{\hskip 10pt}
      *{2}{r}@{}}
    \toprule
    & \multicolumn{2}{c}{$V\!=\!8\text{K}$}
    & \multicolumn{2}{c}{$V\!=\!16\text{K}$}
    & \multicolumn{2}{c}{$V\!=\!32\text{K}$}
    & \multicolumn{2}{c}{$V\!=\!64\text{K}$}
    & \multicolumn{2}{c}{$V\!=\!128\text{K}$}
    & \multicolumn{2}{c}{$V\!=\!256\text{K}$} \\
    \cmidrule(lr){2-3}\cmidrule(lr){4-5}\cmidrule(lr){6-7}
    \cmidrule(lr){8-9}\cmidrule(lr){10-11}\cmidrule(lr){12-13}
    $B$ & bit & adp & bit & adp & bit & adp & bit & adp & bit & adp & bit & adp \\
    \midrule
     1 & \textbf{1.00} & 1.31 & \textbf{1.00} & 1.21 & \textbf{1.00} & 1.24 & \textbf{1.00} & 1.24 & \textbf{1.00} & 1.24 & \textbf{1.00} & 1.14 \\
     2 & \textbf{1.13} & 1.33 & \textbf{1.02} & 1.21 & \textbf{1.00} & 1.24 & \textbf{1.00} & 1.23 & \textbf{1.02} & 1.25 & \textbf{1.00} & 1.12 \\
     4 & \textbf{1.13} & 1.33 & \textbf{1.05} & 1.23 & \textbf{1.01} & 1.26 & \textbf{1.00} & 1.24 & \textbf{1.02} & 1.25 & \textbf{1.12} & 1.14 \\
     8 & \textbf{1.13} & 1.38 & \textbf{1.07} & 1.25 & \textbf{1.03} & 1.26 & \textbf{1.02} & 1.26 & 1.29 & \textbf{1.27} & 1.57 & \textbf{1.46} \\
    16 & \textbf{1.13} & 1.38 & \textbf{1.18} & 1.25 & \textbf{1.17} & 1.29 & 1.39 & \textbf{1.34} & 1.88 & \textbf{1.76} & 2.54 & \textbf{2.49} \\
    32 & \textbf{1.29} & 1.40 & \textbf{1.45} & 1.57 & 1.64 & \textbf{1.56} & \textbf{2.09} & 2.10 & \textbf{3.00} & 3.10 & \textbf{4.53} & \textbf{4.53} \\
    \bottomrule
    \end{tabular}
    \vspace{6pt}
    \caption{%
      Normalized latency of \textsc{bitonic} (bit) vs.\ \textsc{adaptive} (adp) sorting strategies
      across vocab sizes and batch sizes on H100 ($k=128$).
      All values are divided by $\min(\text{bitonic}_1,\,\text{adaptive}_1)$ for
      the corresponding architecture and vocab size, so 1.00 anchors the fastest
      \texttt{batch\_size\,=\,1} configuration.
      \textbf{Bold} marks the lower (better) value per cell; ties bold both.%
    }
\end{table}

\section{Additional Accuracy Results}
\label{app:acc}

\begin{figure}[h]
  \centering
  \quad
  \quad
  \begin{subfigure}[b]{0.35\textwidth}
      \centering
      \includegraphics[height=5.5cm]{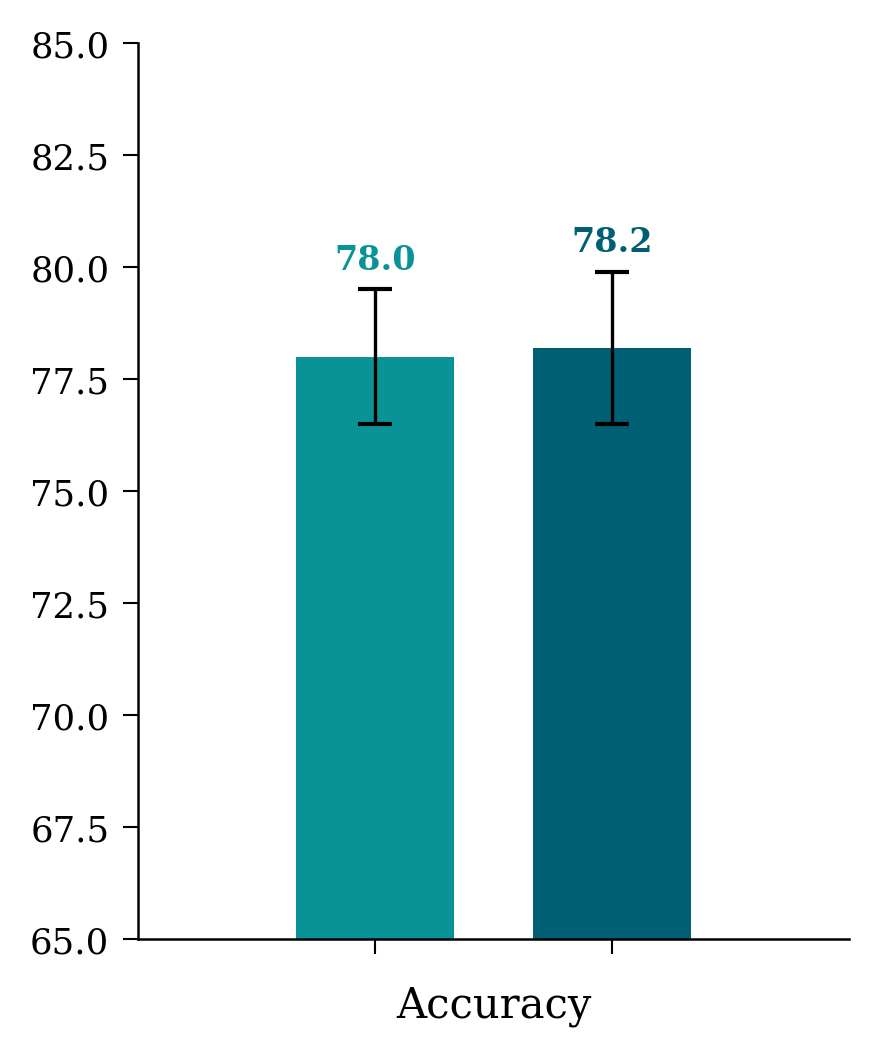}
      \caption{DeepSeek-V3.1 - GPQA}
      \label{fig:gpqa-accuracy}
  \end{subfigure}
  \hfill
  \begin{subfigure}[b]{0.55\textwidth}
      \centering
      \includegraphics[height=5.5cm]{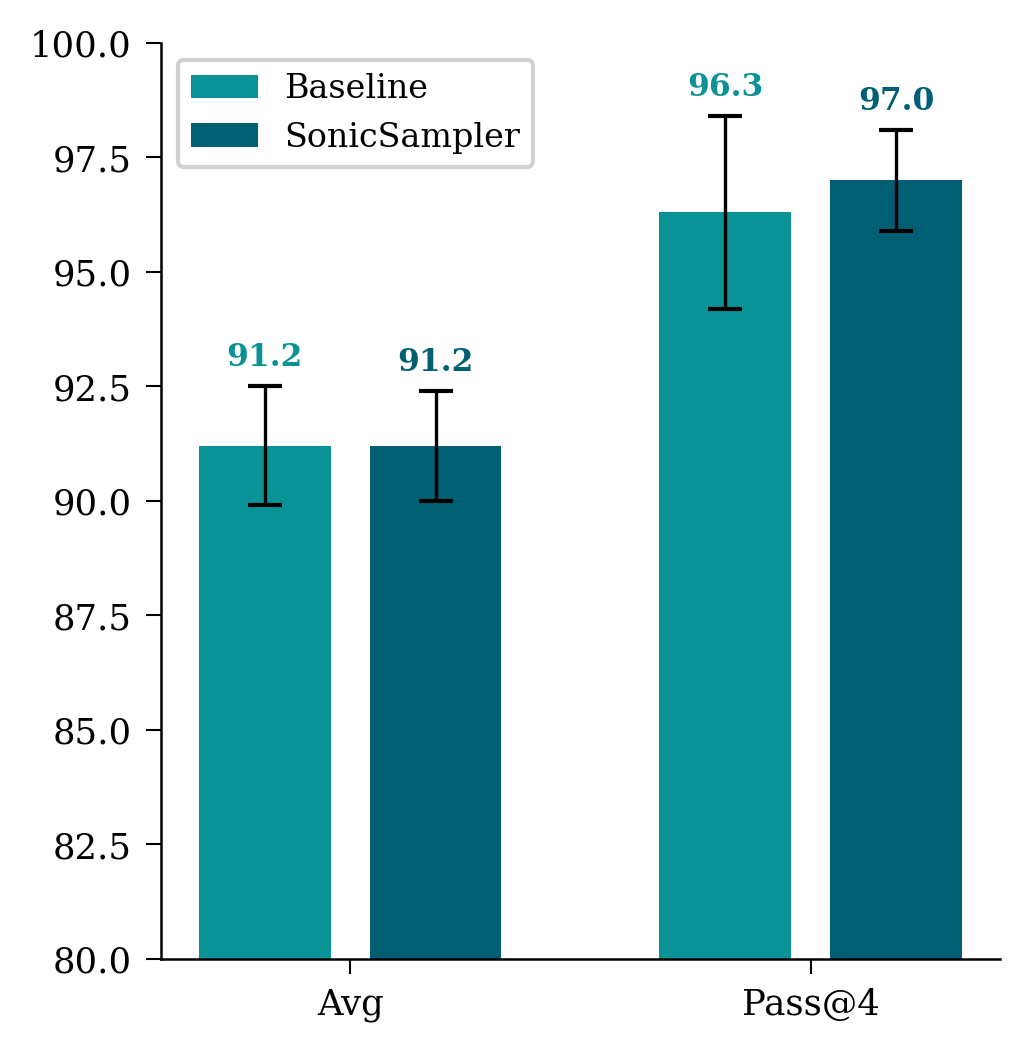}
      \caption{DeepSeek-V3.1 - AIME 2024}
      \label{fig:aime24-accuracy}
  \end{subfigure}
  \quad
  \quad
  \caption{End-to-end accuracy on DeepSeek-V3.1. Error bars show standard deviation over 10 runs.}
  \label{fig:accuracy-comparison}
\end{figure}

\Cref{fig:accuracy-comparison} shows that the introduction of the constrained Top-$k$ setting ($k \leq 128$) in SonicSampler does not degrade the model quality across benchmarks. In GPQA, both configurations achieve nearly identical accuracy (78.0 vs. 78.2), with overlapping error bars that indicate no statistically significant differences.

On AIME 2024, we observe the same trend where average accuracy remains unchanged.
Importantly, these differences fall within the observed variance across runs, suggesting that the constrained sampling space does not negatively impact solution diversity or correctness.

\section{Additional Probability Mass Validation under Bounded Top-$k$}
\label{app:mass_validation}

\begin{figure}[t]
    \centering
    \begin{subfigure}[t]{\linewidth}
        \centering
        \includegraphics[width=0.9\linewidth]{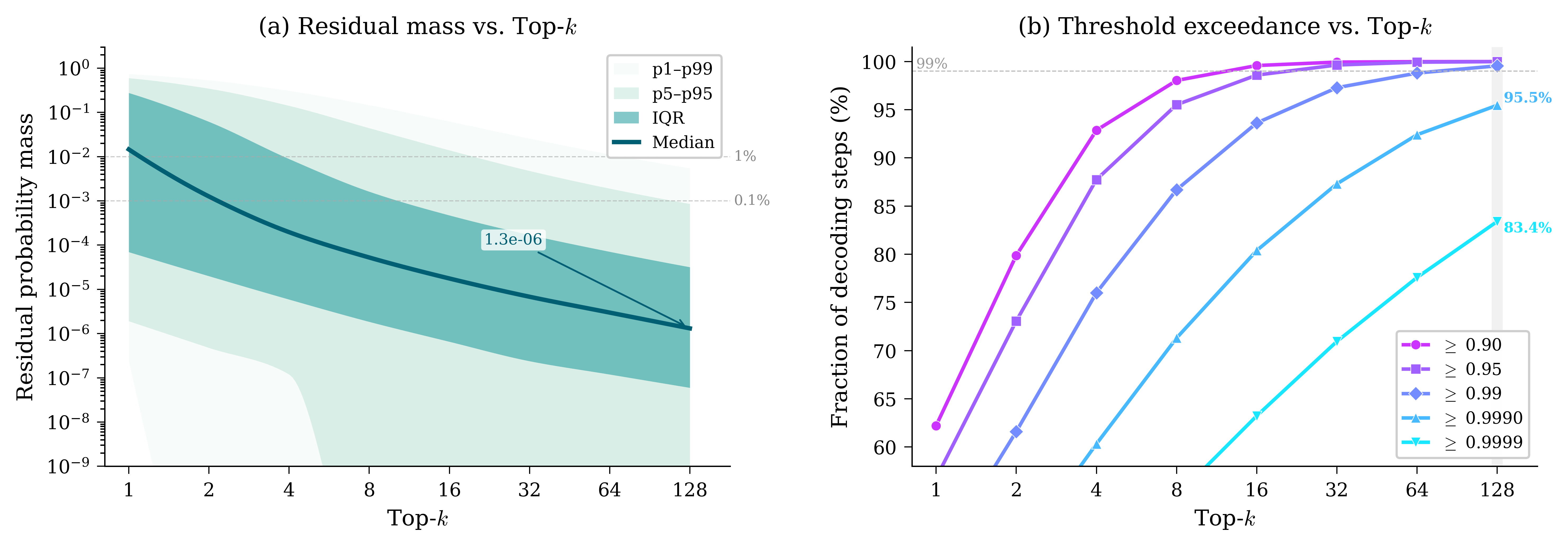}
        \caption{GLM-4.7}
        \label{fig:topk-mass-glm}
    \end{subfigure}

    \vspace{0.8em}

    \begin{subfigure}[t]{\linewidth}
        \centering
        \includegraphics[width=0.9\linewidth]{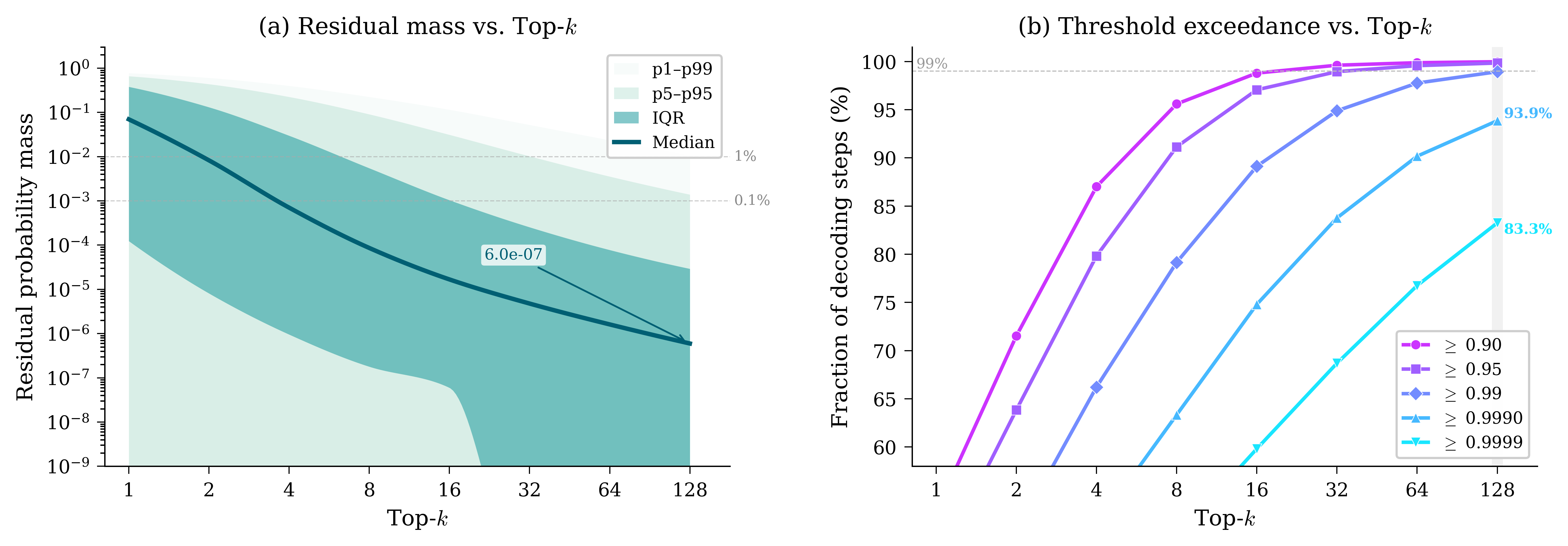}
        \caption{GPT-OSS-120B}
        \label{fig:topk-mass-gptoss}
    \end{subfigure}

    \vspace{0.8em}

    \begin{subfigure}[t]{\linewidth}
        \centering
        \includegraphics[width=0.9\linewidth]{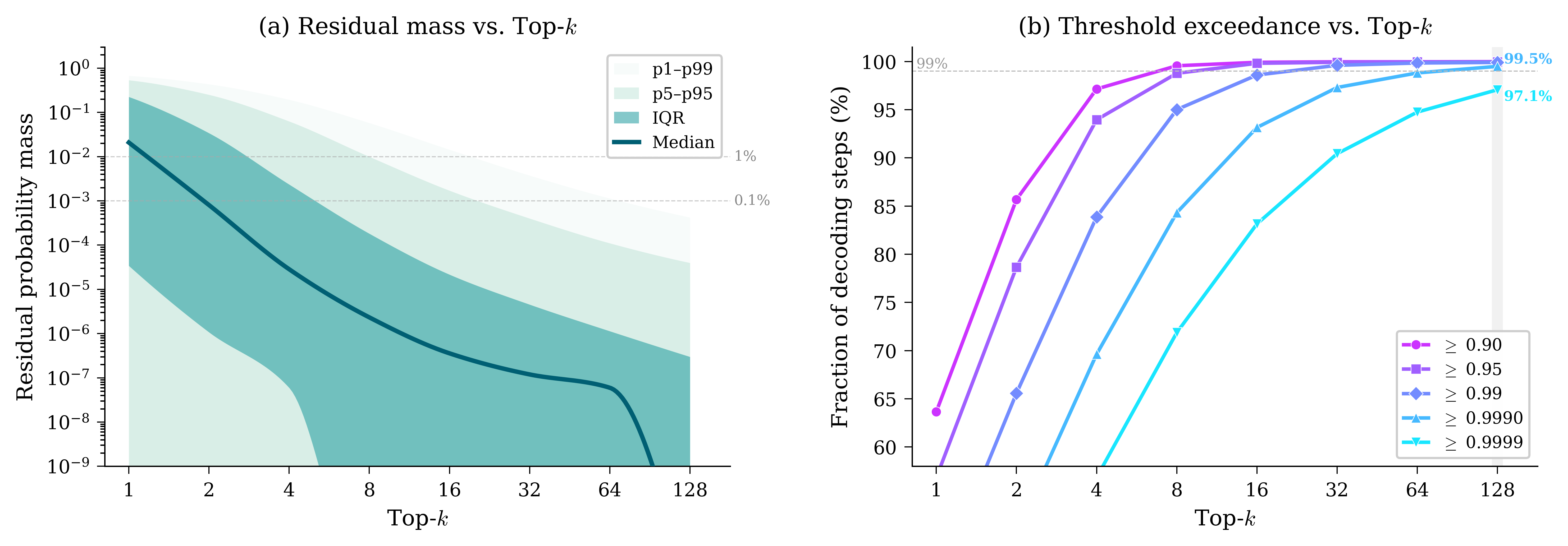}
        \caption{Qwen3-8B}
        \label{fig:topk-mass-qwen}
    \end{subfigure}

    \caption{Residual probability mass outside the Top-$k$ set on additional models. Across all three model families, probability mass decays rapidly with $k$, indicating that bounded Top-$k$ retains the effective support relevant for practical sampling thresholds.}
    \label{fig:topk-mass-extra}
\end{figure}

This section extends the bounded Top-$k$ analysis from \Cref{exp:correctness} to additional models. In the main text, we show on DeepSeek v3.1 that the residual probability mass outside the retained Top-$k$ set decays rapidly, making bounded Top-$k$ effectively lossless in practice. \Cref{fig:topk-mass-extra} shows that the same qualitative behavior holds across GLM-4.7~\citep{5team2025glm45agenticreasoningcoding}, GPT-OSS-120B~\citep{openai2025gptoss120bgptoss20bmodel}, and Qwen3-8B~\citep{yang2025qwen3technicalreport}. Across these models, probability mass concentrates quickly in a small candidate set, providing further empirical support for using $k=128$ in SonicSampler.

\end{document}